\documentclass[10pt,twocolumn,letterpaper]{article}

\usepackage{cvpr}              % To produce the CAMERA-READY version
\definecolor{cvprblue}{rgb}{0.21,0.49,0.74}
\usepackage[pagebackref,breaklinks,colorlinks,allcolors=cvprblue]{hyperref}
\usepackage{multirow}
\usepackage{booktabs}
\usepackage[table,xcdraw]{xcolor}

\definecolor{color_blue}{RGB}{252,182,165}
\definecolor{color_pink}{RGB}{255,217,178}
\definecolor{color_yellow}{RGB}{255,255,204}
\definecolor{color_blue1}{RGB}{135,206,235}

\def\paperID{} % *** Enter the Paper ID here
\def\confName{CVPR}
\def\confYear{2026}

\title{Negative Binomial Variational Autoencoders for Overdispersed Latent Modeling}

\author{Yixuan Zhang\textsuperscript{\rm 1}\thanks{Equal contribution.}, Jinhao Sheng\textsuperscript{\rm 2}\footnotemark[1], Wenxin Zhang\textsuperscript{\rm 3}, Quyu Kong\textsuperscript{\rm 4}, Feng Zhou\textsuperscript{\rm 5}\thanks{Corresponding author.}\\
\textsuperscript{\rm 1}Southeast University, \textsuperscript{\rm 2}China Medical University Shenyang, \\\textsuperscript{\rm 3}University of Chinese Academy of Science, \textsuperscript{\rm 4}Alibaba Cloud, \\
\textsuperscript{\rm 5} Center for Applied Statistics and School of Statistics, Renmin University of China\\
{\tt\small zh1xuan@hotmail.com, jhsheng@cmu.edu.cn, feng.zhou@ruc.edu.cn}
}

\begin{document}
\maketitle
\begin{abstract}
Although artificial neural networks are often described as brain-inspired, their representations typically rely on continuous activations, such as the continuous latent variables in variational autoencoders (VAEs), which limits their biological plausibility compared to the discrete spike-based signaling in real neurons. Extensions like the Poisson VAE introduce discrete count-based latents, but their equal mean-variance assumption fails to capture overdispersion in neural spikes, leading to less expressive and informative representations. To address this, we propose NegBio-VAE, a negative-binomial latent-variable model with a dispersion parameter for flexible spike count modeling. NegBio-VAE preserves interpretability while improving representation quality and training feasibility via novel KL estimation and reparameterization. Experiments on four datasets demonstrate that NegBio-VAE consistently achieves superior reconstruction and generation performance {compared to competing single-layer VAE baselines}, and yields robust, informative latent representations for downstream tasks. 
Extensive ablation studies are performed to verify the model’s robustness w.r.t. various components. Our code is available at \url{https://github.com/co234/NegBio-VAE}.
\end{abstract}    
\section{Introduction}
\label{sec:intro}
Although artificial neural networks (ANNs) have historically been described as brain-inspired, their design choices are primarily driven by computational considerations rather than strict biological fidelity~\cite{tavanaei2019deep,arbib2003handbook}. A key distinction lies in how information is represented: while biological neurons communicate through sequences of action potentials (spike trains)~\cite{perkel1967neuronal}, most machine learning models adopt continuous activations. This contrast has motivated a line of work that investigates discrete, spike like representations as a pathway toward enriching the expressiveness of generative models~\cite{mainen1995reliability,bair1996temporal,gollisch2008rapid}. From this perspective, studying count-based representations is not only biologically inspired but also methodologically valuable for expanding the modeling capacity of deep generative frameworks~\cite{van2020brain,ghosh2009spiking}.

Among these frameworks, the variational autoencoder (VAE)~\cite{kingma2013auto} is a powerful generative model grounded in Bayesian inference that learns structured latent representations of data, and is often described as brain-inspired due to its similarity to how the brain encodes sensory information~\cite{marino2022predictive,vafaii2023hierarchical,storrs2021unsupervised}.
While VAEs have achieved broad success, they typically employ continuous latent variables, in contrast to the discrete spike counts encoded by the brain. To bridge this gap, recent works have proposed extensions such as categorical or Poisson VAEs~\cite{jang2017categorical,van2017neural,vafaii2024poisson}, which introduce discrete latent variables that not only offer greater biological plausibility but also enhance the capacity to model categorical or count structures in latent variables. 

The main improvement presented in this paper builds on the Poisson VAE ($\mathcal{P}$-VAE)~\cite{vafaii2024poisson}, which encodes data as discrete spike counts drawn from a Poisson distribution. While the Poisson model provides a natural starting point, it imposes a restrictive assumption: the mean and variance of the discrete spike counts must be equal. In practice, however, neural spike trains often exhibit overdispersion, where the variance of the spike counts significantly exceeds the mean~\cite{taouali2016testing,moshitch2014using,stevenson2016flexible}. {This has been linked to neurobiological sources such as trial-to-trial gain variability and network-level fluctuations~\cite{stevenson2016flexible}. While underdispersion can arise in neurons with refractory periods~\cite{berry1997structure}, overdispersion is the more prevalent and consequential deviation from Poisson statistics across cortical recordings~\cite{shadlen1998variable,goris2014partitioning}.} This coupling of the mean and variance limits the flexibility of the latent space, leading to underestimated uncertainty and reduced representational expressiveness. 

%causing the model to underestimate uncertainty and fail to capture overdispersed structures commonly found in neural spikes, thereby limiting the expressive power of the learned representations. 

% \begin{figure*}[t]
%   \centering
%   \includegraphics[width=\linewidth]{plots/recons/mnisthightlight.pdf}
%   \caption{Reconstructions from different VAEs (256 dims) on MNIST. Despite using the same latent dimensionality, NegBio-VAE recovers sharper digit structures (e.g., the gap in \textbf{0} and the loop in \textbf{4}) that others miss. This advantage stems from its ability to model overdispersed latent spike counts, providing greater variance flexibility and enabling sharper distinctions in high-variability regions.}
%   \label{fig:mnist-highlight}
% \end{figure*}

\begin{figure*}[!t]
\centering
\includegraphics[width=\linewidth]{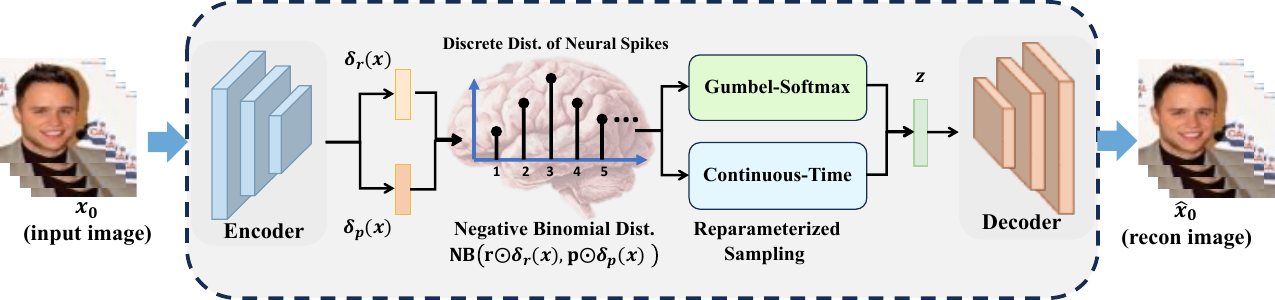}
\caption{Overview of the proposed NegBio-VAE framework. The data are encoded as discrete spike counts drawn from a negative binomial distribution, whose variance exceeds the mean, enabling the model to capture overdispersed latent structures.}
\label{fig: method overview}
\end{figure*}

To address this limitation, we adopt the negative binomial (NB) distribution~\cite{ross1985negative}, a two-parameter generalization of the Poisson distribution that introduces a dispersion parameter, allowing the variance to exceed the mean. 
This flexibility allows modeling of overdispersed spike counts, enabling latent representations that better capture the heterogeneous variability. Building on this idea, we propose NegBio-VAE (see \cref{fig: method overview}), a principled extension of the VAE framework that preserves count-based representations while more accurately reflecting their statistical variability. While this formulation greatly enhances representational flexibility, it also introduces two challenges: (1) computing the KL divergence between NB distributions, and (2) performing reparameterized sampling. We address both with efficient approximations that make NegBio-VAE practically trainable. 
Empirically, NegBio-VAE demonstrates superior reconstruction quality, stronger generative performance, and more informative latent representations for downstream tasks.

% Our main contributions can be summarized as follows:
% \textbf{(1)} We propose NegBio-VAE, which introduces an additional dispersion parameter to model overdispersed latent spike counts, substantially enhancing the flexibility and expressiveness of latent representations. \textbf{(2)} We develop efficient training techniques, including two KL estimation strategies (Monte Carlo and dispersion sharing) and two differentiable reparameterization methods (Gumbel–Softmax relaxation and continuous-time simulation), enabling stable gradient-based optimization. 
% \textbf{(3)} Extensive experiments on four benchmark datasets show that NegBio-VAE consistently outperforms strong baselines in both reconstruction and generation, while yielding more informative latent representations for downstream tasks. Comprehensive ablations further verify its robustness.
Our main contributions are summarized as follows:
\textbf{(1)} We propose NegBio-VAE, which introduces a dispersion parameter to model overdispersed latent spike counts and improve the flexibility of latent representations.
\textbf{(2)} We develop efficient training strategies with two KL estimators (Monte Carlo and Dispersion Sharing) and two differentiable reparameterizations (Gumbel–Softmax and Continuous-time Simulation) for stable optimization.
\textbf{(3)} Experiments on four benchmark datasets show that NegBio-VAE outperforms strong baselines in reconstruction and generation while learning more informative latent representations for downstream tasks.

\section{Related Works}
\label{sec:related_work}
\textbf{Brain-like ANNs}, emerging at the intersection of neuroscience and machine learning, aim to mirror the brain’s functionality and structure. Related works can be categorized into two types: spiking neural networks (SNNs) and brain-like generative models. SNNs~\citep{ghosh2009spiking,cheng2020lisnn,zheng2021going,fang2021incorporating,li2021differentiable}, like biological neurons, use discrete spikes for communication instead of continuous activations as in traditional ANNs. A notable model is the leaky integrate-and-fire (LIF) model, which simulates the temporal dynamics of spike generation. The second category includes generative models that learn data representations similar to how brain processes sensory information. Key works in this area include brain-like VAEs~\citep{kamata2022fully,yadav2025differentially,vafaii2024poisson}, GANs~\citep{kotariya2022spiking,rosenfeld2022spiking,feng2024spiking}, and diffusion models~\citep{liu2024spiking,cao2024spiking,kapoor2024latent}.
Our work extends the $\mathcal{P}$-VAE~\citep{vafaii2024poisson} by incorporating a NB distribution to better capture overdispersion in latent spike counts, enabling richer and more flexible variability in the latent representations. 

\textbf{Discrete VAEs} are typically categorized into two types: discrete representations and discrete data. In VAEs with discrete representations, the variables capture the underlying discrete structure of the data. Most current works on discrete-representation VAEs use categorical distributions for the latent variables~\citep{van2017neural,fortuinsom,jang2017categorical,dupont2018learning}. Other works employ Bernoulli~\citep{kamata2022fully,rolfe2017discrete} or Poisson distributions~\citep{vafaii2024poisson,zhan2023esvae}. These methods have achieved significant success in speech synthesis and image generation. The second category focuses on VAEs for discrete data, such as text, categorical, or count data. These models reconstruct discrete data, making them suitable for tasks like natural language processing and structured prediction~\citep{zhao2020variational,polykovskiy2020deterministic}. While \citep{zhao2020variational} uses the NB distribution to model count data while keeping the latent variables continuous, our work extends NB modeling to discrete latent variables in a VAE. 

\section{Preliminaries}
\label{sec: preliminaries}
This section reviews VAE and $\mathcal{P}$-VAE, first covering the standard VAE framework and then its adaptation to model latent spike counts with a Poisson distribution.

\subsection{Variational Autoencoder} 
% VAE~\cite{kingma2014autoencoding} is a probabilistic generative model that defines a joint distribution $p(\mathbf{x},\mathbf{z})$ over observed data $\mathbf{x}$ and latent variables $\mathbf{z}$. Data are generated by sampling $\mathbf{z} \sim p(\mathbf{z})$ (the prior) and decoding it via $p_\theta(\mathbf{x}\mid \mathbf{z})$. Latents are inferred using an approximate posterior $q_\phi(\mathbf{z}\mid \mathbf{x})$, which estimates the true posterior distribution and is known as the encoder. Parameters are learned by maximizing the evidence lower bound (ELBO), a tractable surrogate for $\log p(\mathbf{x})$. The ELBO objective encourages accurate reconstruction of the input data while regularizing the latent space to learn meaningful representations, and is usually given as:  
VAE~\cite{kingma2014autoencoding} is a probabilistic generative model defining a joint distribution $p(\mathbf{x},\mathbf{z})$ over data $\mathbf{x}$ and latent variables $\mathbf{z}$. Samples are generated by $\mathbf{z}\sim p(\mathbf{z})$ and decoded via $p_\theta(\mathbf{x}\mid\mathbf{z})$, while inference uses an approximate posterior $q_\phi(\mathbf{z}\mid\mathbf{x})$. Model parameters are learned by maximizing the evidence lower bound (ELBO), a tractable surrogate of $\log p(\mathbf{x})$ that balances reconstruction and latent regularization:
\begin{equation*}
\label{eq:vae_loss}
    \mathcal{L}_{\text{VAE}} = \mathbb{E}_{q_\phi(\mathbf{z}\mid \mathbf{x})}[\log p_\theta(\mathbf{x}\mid \mathbf{z})] - \mathcal{D}_{\text{KL}}[q_\phi(\mathbf{z}\mid \mathbf{x}) || p(\mathbf{z})].
\end{equation*}
The first term enforces faithful reconstruction, while the second term regularizes the latent space. VAEs enable gradient-based optimization via the reparameterization trick~\cite{kingma2014autoencoding}, which introduces differentiable sampling between the encoder and decoder. Standard implementations assume an isotropic Gaussian prior $p(\mathbf{z})$, simplifying computation but limiting expressiveness.

%To enable gradient-based optimization, VAEs employ the reparameterization trick~\cite{kingma2014autoencoding}, introducing a differentiable sampling step between the encoder and decoder. In standard implementations, {the prior $p(\mathbf{z})$ is modeled as an isotropic Gaussian,} which simplifies computation but limits expressiveness in certain settings.

\subsection{Poisson VAE} 
To better mimic biological neuron activity, the $\mathcal{P}$-VAE~\cite{vafaii2024poisson} was proposed to model spike counts as discrete latent variables. Specifically, it uses the Poisson distribution to represent the spike counts of $K$ neurons, with the latent variable $\mathbf{z} \in \mathbb{Z}_0^{+K}$. The prior and variational posterior are defined as: 
\begin{equation*}
\begin{aligned}
    \text{Prior: } \quad p(\mathbf{z}) &= \text{Poi}(\mathbf{z};\mathbf{r}), \quad \quad \\ \text{Posterior: } \quad q(\mathbf{z}\mid \mathbf{x}) &= \text{Poi}(\mathbf{z};\mathbf{r}\odot \boldsymbol{\delta}_r(\mathbf{x})), 
\end{aligned}
\end{equation*}
where both the prior Poisson and the posterior Poisson are factorized, i.e., $\text{Poi}(\mathbf{z}) = \prod_{i=1}^K \text{Poi}(z_i)$. Here, $\mathbf{r} \in \mathbb{R}^{+K}$ denotes the prior firing rates, and $\mathbf{r} \odot \boldsymbol{\delta}_r(\mathbf{x})$ gives the posterior firing rates, with $\odot$ denoting element-wise multiplication. The encoder output $\boldsymbol{\delta}_r(\mathbf{x}) \in \mathbb{R}^{+K}$ modulates the ratio of posterior to prior firing rates based on the input. In contrast to standard VAEs where latent variables are continuous and typically drawn from a Gaussian, the $\mathcal{P}$-VAE models $\mathbf{z}$ as a vector of discrete spike counts, which better resembles neural firing behavior. The objective of $\mathcal{P}$-VAE is given by: 
\begin{equation}
    \mathcal{L}_{\mathcal{P}\text{-VAE}} = \mathbb{E}_{\text{Poi}(\mathbf{z};\mathbf{r}\odot \boldsymbol{\delta}_r(\mathbf{x}))}\left [\log p_\theta(\mathbf{x}\mid \mathbf{z})\right ] + \sum^K_{i=1}r_i g(\delta_{r_i}),
\end{equation}
where $g(a)=1-a+a\log a$ corresponds to the KL divergence between two Poisson distributions. 
\section{Methodology}
\label{sec: methods}
A key limitation of the Poisson distribution is its restrictive assumption that the mean and variance of spike counts are equal. This assumption fails to capture the overdispersion frequently observed in neural spike train. To address this, we propose the NegBio-VAE, which applies a more flexible NB distribution. As a two-parameter generalization of the Poisson, the NB distribution introduces a dispersion parameter that allows the variance to exceed the mean. This makes it more suitable for modeling overdispersed spike counts. The NB distribution has been widely applied in various fields, such as spiking neuron models~\cite{pillow2012fully}, RNA sequence analysis~\cite{di2011nbp}, and language modeling~\cite{zhou2013negative}.

We begin by defining the prior and posterior distributions over the latent spike counts $\mathbf{z} \in \mathbb{Z}_0^{+K}$ as $p(\mathbf{z}) = \text{NB}(\mathbf{z};\mathbf{r},\mathbf{p})$ and $q(\mathbf{z}\mid \mathbf{x}) =\text{NB}(\mathbf{z};\mathbf{r} \odot \boldsymbol{\delta}_r(\mathbf{x}),\mathbf{p} \odot \boldsymbol{\delta}_p(\mathbf{x}))$, respectively. 
Similar to $\mathcal{P}$-VAE, both the prior and posterior NB distribution are factorized, i.e., $\text{NB}(\mathbf{z}) = \prod_{i=1}^K \text{NB}(z_i)$,  $\boldsymbol{\delta}_r(\mathbf{x})$ and $\boldsymbol{\delta}_p(\mathbf{x})$ are outputs of the encoder, which captures the ratio of the posterior parameters to the prior parameters. With this setup, the ELBO of NegBio-VAE becomes: 
\begin{equation}
\begin{aligned}
    \mathcal{L} &= \mathbb{E}_{\text{NB}(\mathbf{z};\mathbf{r} \odot \boldsymbol{\delta}_r(\mathbf{x}),\mathbf{p} \odot \boldsymbol{\delta}_p(\mathbf{x}))}\left[ \log p_\theta(\mathbf{x}\mid \mathbf{z}) \right]\\ &- \mathcal{D}_{\text{KL}}[\text{NB}(\mathbf{z};\mathbf{r} \odot \boldsymbol{\delta}_r(\mathbf{x}),\mathbf{p} \odot \boldsymbol{\delta}_p(\mathbf{x})) || \text{NB}(\mathbf{z};\mathbf{r},\mathbf{p})].
\label{nbvae_loss}
\end{aligned}
\end{equation}
While this formulation enables greater flexibility, it also introduces two key technical challenges during the training of NegBio-VAE: (1) The second term in \cref{nbvae_loss} requires calculating the KL divergence between two NB distributions; (2) The first term in \cref{nbvae_loss} requires reparameterized sampling from the NB distribution. We address each of these issues in the following sections. 

\subsection{KL Divergence between NB Distributions}
\label{sec:method-kl-divergence}
In both vanilla VAE and $\mathcal{P}$-VAE, the KL term is tractable due to closed-form solutions for Gaussian and Poisson distributions. However, no such form exists for the KL divergence between two NB distributions, which poses the first challenge for training NegBio-VAE. To address this, we propose two strategies: a \textbf{Monte Carlo} method for direct approximation, and a \textbf{dispersion sharing} technique that simplifies the KL divergence by partially tying posterior parameters to the prior. 

\textbf{(1) Monte Carlo.} 
% Optimizing the ELBO for NegBio-VAE is challenging due to the absence of an analytical solution for the KL divergence between two NB distributions. This prevents direct computation of the KL term in the objective function. However, we can overcome this issue by resorting to Monte Carlo estimation. Using the identity $\mathcal{D}_{\text{KL}}[q(\mathbf{z}) || p(\mathbf{z})] = \mathbb{E}_{q(\mathbf{z})}\left[ \log q(\mathbf{z})-\log p(\mathbf{z}) \right]$, we can approximate the KL divergence by Monte Carlo sampling from the variational posterior and computing the sample average of the log-density difference between the posterior and the prior. 
Optimizing the ELBO for NegBio-VAE is challenging due to the lack of an analytical form for the KL divergence between two NB distributions, preventing direct computation of the KL term. We address this using Monte Carlo estimation. Specifically, using $\mathcal{D}{\text{KL}}[q(\mathbf{z}) || p(\mathbf{z})] = \mathbb{E}{q(\mathbf{z})}\left[\log q(\mathbf{z})-\log p(\mathbf{z})\right]$, the KL divergence is approximated by sampling from the variational posterior and averaging the log-density difference between posterior and prior.

Substituting this expression into the ELBO in \cref{nbvae_loss} we obtain the following objective: 
\begin{equation*}
\begin{aligned}
    \mathcal{L} &= \mathbb{E}_{\text{NB}(\mathbf{z};\mathbf{r} \odot \boldsymbol{\delta}_r(\mathbf{x}),\mathbf{p} \odot \boldsymbol{\delta}_p(\mathbf{x}))}[ 
    \log p_\theta(\mathbf{x}\mid \mathbf{z}) \\ & - \log \text{NB}(\mathbf{z};\mathbf{r} \odot \boldsymbol{\delta}_r(\mathbf{x}),\mathbf{p} \odot \boldsymbol{\delta}_p(\mathbf{x})) + \log \text{NB}(\mathbf{z};\mathbf{r},\mathbf{p})]. 
\end{aligned}
\end{equation*}
Clearly, as long as we can implement reparameterized sampling from the NB distribution, we can use the above objective function to train NegBio-VAE. 

\textbf{(2) Dispersion Sharing.} 
Although the KL divergence between two general NB distributions, $\text{NB}(z;r_1,p_1)$ and $\text{NB}(z;r_2,p_2)$, does not have an analytical solution, a tractable analytical form exists when the dispersion parameters are shared, i.e., $r_1 = r_2 = r$. 

Based on this observation, we propose an alternative strategy for computing the KL term in the NegBio-VAE by constraining the prior $\text{NB}(\mathbf{z};\mathbf{r},\mathbf{p})$ and the posterior $\text{NB}(\mathbf{z};\mathbf{r} \odot \boldsymbol{\delta}_r(\mathbf{x}), \mathbf{p} \odot \boldsymbol{\delta}_p(\mathbf{x}))$ to share the same dispersion parameter, i.e., setting $\boldsymbol{\delta}_r(\mathbf{x})$ to be $\mathbf{1}$. Then, the KL term in \cref{nbvae_loss} admits a closed-form solution: 
\begin{equation}
    \mathcal{D}_{\text{KL}}[\text{NB}(\mathbf{z};\mathbf{r},\mathbf{p} \odot \boldsymbol{\delta}_p(\mathbf{x})) || \text{NB}(\mathbf{z};\mathbf{r},\mathbf{p})] = \sum^K_{i=1}r_i g(p_i,\delta_{p_i}), 
\end{equation}
where $g(a,b)$ is defined as: $g(a,b) = \log b + \frac{1-ab}{ab} \log \left(\frac{1-ab}{1-a}\right)$, with $a\in (0,1)$ and $b>0$. The complete derivation can be found in appendix. Then, the final NegBio-VAE objective becomes: 
\begin{equation}
    \mathcal{L} = \mathbb{E}_{\text{NB}(\mathbf{z};\mathbf{r},\mathbf{p} \odot \boldsymbol{\delta}_p(\mathbf{x}))}\left [ \log p_\theta(\mathbf{x}\mid \mathbf{z}) \right ]+  \sum^K_{i=1}r_i g(p_i,\delta_{p_i}). 
\end{equation}
Importantly, sharing the same dispersion parameter between the prior and posterior does not imply that they have identical means or variances. For the NB distribution, the mean is given by $r(1-p)/p$ and the variance by $r(1-p)/p^2$. Thus, even when $r$ is the same for both, different $p$ still allows the posterior to capture different distributional properties from the prior. 

Both methods have advantages and limitations. The Monte Carlo method makes no assumptions about the variational posterior but may yield higher-variance gradient estimates. The dispersion sharing method instead assumes a shared dispersion parameter, enabling analytic KL computation. Although analytic KL does not guarantee lower gradient variance, it simplifies optimization and often improves training stability in practice while preserving the ability to capture overdispersion.
%The Monte Carlo approach makes no assumptions about the variational posterior but suffers from higher gradient variance, which may lead to unstable training. In contrast, the dispersion sharing method assumes a shared dispersion parameter for the variational posterior, enabling analytical computation of the KL divergence and reducing gradient variance. Importantly, this assumption does not compromise the model’s ability to capture overdispersion, providing a practical trade-off between expressive power and training stability. 
We compare the performance of both strategies, and the results are presented in \cref{sec: exp}.

\subsection{Reparameterized Sampling for NB Distribution}
The second challenge in training NegBio-VAE lies in the sampling process. The expectation term in the ELBO requires reparameterized sampling from the NB distribution to allow efficient gradient-based optimization. 
Reparameterizing discrete distributions is more challenging compared to continuous ones, but it can be achieved through suitable relaxation techniques. 
In this section, we describe how to apply reparameterization to the NB distribution by leveraging a key property: the NB distribution can be represented as a continuous mixture of Poisson distributions, where the mixing weight being a Gamma distribution: 
\begin{equation}
    \text{NB}(z;r,p)=\int_0^\infty \text{Poi}(z|\lambda)\text{Gamma}(\lambda;r,\frac{p}{1-p}) d\lambda. 
\end{equation}
This implies that a sample from $\text{NB}(z;r,p)$ can be obtained by first sampling $\lambda \sim \text{Gamma}(r, \frac{p}{1-p})$, followed by sampling $z \sim \text{Poi}(\lambda)$. 

The first step, sampling from the Gamma distribution, is straightforward to reparameterize {via implicit reparameterization gradients~\cite{NEURIPS2018_92c8c96e}}. In practice, PyTorch's \verb|Gamma.rsample()| function supports gradient propagation, as it uses the Marsaglia-Tsang algorithm in its underlying implementation and ensures differentiability through implicit gradient computation. The second step, sampling from the Poisson distribution, is more challenging, as it lacks a standard reparameterizable form. To address this, we adopt approximate relaxation techniques such as \textbf{Gumbel-Softmax Relaxation}~\cite{jang2017categorical} and \textbf{Continuous-Time Simulation}~\cite{vafaii2024poisson}. Both methods rely on a temperature parameter to transform ``hard'' counts into ``soft'' counts, thereby enabling differentiability. For implementation details, please see appendix.

\textbf{(1) Gumbel-Softmax Relaxation.} 
To enable differentiable sampling from a Poisson distribution, we adopt a relaxation-based strategy that treats the Poisson as a categorical distribution over a truncated support $ \{ 0, 1, \ldots, Z_{\text{max}} \}$. By using the Gumbel-Softmax trick~\cite{jang2017categorical}, we construct a soft approximation of the discrete counts: 
\begin{equation*}
    \tilde{z} = \sum_{z=0}^{Z_{\text{max}}} z \cdot \mathrm{softmax}\left( \frac{\log \text{Poi}(z) + \epsilon_z}{\tau} \right),
\end{equation*}
where $\epsilon_z \sim \text{Gumbel}(0, 1)$ is an i.i.d. Gumbel noise and $\tau >0$ is a temperature controlling the degree of relaxation. As $\tau \xrightarrow{} 0$, the soft sample $\tilde{z}$ converges to the Poisson distribution. 

\textbf{(2) Continuous-Time Simulation.}
{Following ~\citet{vafaii2024poisson}, we adopt the continuous-time simulation method}, which leverages the connection between the Poisson distribution and the Poisson process. It models a Poisson-distributed count as the number of events occurring within the interval $[0,1]$, where inter-arrival times follow an exponential distribution with rate $\lambda$. The soft count is computed by simulating the inter-arrival times and accumulating a temperature-smoothed approximation of the total event count: 
\begin{equation*}
    \tilde{z} = \sum_{n=1}^M\sigma\left(\frac{1-S_n}{\tau}\right), 
\end{equation*}
where $S_n = \sum_{i=1}^n s_i, \quad 1\leq n \leq M,  \quad \{s_i\}_{i=1}^M \sim \text{Exponential}(\lambda)$ and $\sigma(\cdot)$ is the sigmoid function, $\tau >0$ is a temperature, and $\tau \xrightarrow{} 0$ converges to the Poisson distribution. This approach enables differentiable Poisson sampling through reparameterizable exponential sampling. 

Both Gumbel-Softmax and continuous-time relaxations are used for the Poisson step in the NB reparameterization. Theoretically, both approaches are valid. Empirically, under the same temperature, we find that the continuous-time relaxation tends to produce smoother count samples, whereas Gumbel-Softmax yields sharper ones. A detailed comparison of the two methods is provided in \cref{sec: exp}. 
% \begin{figure}
%     \centering
%     \includegraphics[width=0.6\linewidth]{plots/reparam_sampling.pdf}
%     \caption{Empirical distributions of Negative Binomial samples generated using Continuous-Time Simulation and Gumbel-Softmax Relaxation. Both methods successfully approximate count-valued outputs consistent with NB sampling behavior, validating their use as differentiable reparameterization strategies. Each method generates 1000 samples using parameters $r = 20$, $p = 0.5$, and temperature $\tau = 0.1$.}
%     \label{fig:reparam_sampling}
% \end{figure}

\begin{table*}[t]\scriptsize
\centering
\caption{Reconstruction and generation performance results on four benchmark datasets. The best and second-best results are marked in \textbf{bold} and \underline{underlined}, respectively.}
\label{tab: main result}
\resizebox{0.86\textwidth}{!}{
\begin{tabular}{c c c c c c c}
\toprule
\multirow{2}{*}{Dataset} & \multirow{2}{*}{Model} & \multicolumn{2}{c}{\cellcolor{color_blue}Reconstruction}  & \multicolumn{3}{c}{\cellcolor{color_pink}Generation} \\
% \cmidrule(lr){3-4} 
% \cmidrule(lr){5-7}
 &  &   \cellcolor{color_blue}MSE  $\downarrow$ & \cellcolor{color_blue}SSIM$\uparrow 
$  & \cellcolor{color_pink}FID@5k $\downarrow$ &  \cellcolor{color_pink}FID@10k $\downarrow$ &\cellcolor{color_pink}KID $\downarrow$  \\
\midrule
\multirow{10}{*}{MNIST} 
 & $\mathcal{G}$-VAE   & 0.0377 & 0.6790 &152.5109 &152.8226 & 0.1788$_{\pm0.0115}$ \\
 & $\mathcal{L}$-VAE  & 0.0377 & 0.7124 &132.7655 &131.7514 & 0.1484$_{\pm0.0103}$\\
 & $\mathcal{C}$-VAE  & 0.0222 & 0.7712 &135.4452 &133.4826 &0.1140$_{\pm0.0132}$\\ 
  & $\mathcal{P}$-VAE   & \underline{0.0125} & \underline{0.8581} & 105.3678 & 104.1416 &0.1250$_{\pm0.0019}$ \\ 
 % & $\text{NVAE}$ &0.0362 &0.7372 &218.1144 &216.3244 &0.2840$_{\pm0.0039}$ \\
 \cmidrule{2-7}
  & \cellcolor{gray!20}$\text{NegBio-VAE}_\text{MC-G}$ & \cellcolor{gray!20}0.0156 & \cellcolor{gray!20}0.8487 & \cellcolor{gray!20}\textbf{79.6727} & \cellcolor{gray!20}\textbf{78.3802} & \cellcolor{gray!20}\textbf{0.0892$_{\pm0.0106}$}\\
 & \cellcolor{gray!20}$\text{NegBio-VAE}_\text{MC-C}$ & \cellcolor{gray!20}\textbf{0.0123} & \cellcolor{gray!20}\textbf{0.8661} & \cellcolor{gray!20}\underline{84.3853} & \cellcolor{gray!20}\underline{83.0010} & \cellcolor{gray!20}\underline{0.0906}$_{\pm0.0111}$\\
 & \cellcolor{gray!20}$\text{NegBio-VAE}_\text{DS-G}$ & \cellcolor{gray!20}0.0168 & \cellcolor{gray!20}0.7960 & \cellcolor{gray!20}87.6456 & \cellcolor{gray!20}87.4101 &\cellcolor{gray!20}0.1000$_{\pm0.0123}$\\
 & \cellcolor{gray!20}$\text{NegBio-VAE}_\text{DS-C}$ & \cellcolor{gray!20}\underline{0.0125} & \cellcolor{gray!20}0.8554 & \cellcolor{gray!20}106.4104 & \cellcolor{gray!20}105.4089 &\cellcolor{gray!20}0.1167$_{\pm0.0094}$\\
\midrule
\multirow{10}{*}{Fashion-MNIST} 
 & $\mathcal{G}$-VAE   &0.1417 &0.1731 &179.8126 & 179.2981 &0.1828$_{\pm0.0106}$ \\
 & $\mathcal{L}$-VAE  & 0.1274 &0.2085 &181.4542 &179.5956 &0.1847$_{\pm0.0112}$  \\
 & $\mathcal{C}$-VAE  & 0.0238 &0.6390 &195.3205 &193.0972 & 0.1835$_{\pm0.0219}$
 \\
 & $\mathcal{P}$-VAE   & \underline{0.0145} &\underline{0.7387} &145.9776 & 146.0128 & 0.1667$_{\pm0.0133}$  \\
 % & $\text{NVAE}$ &0.1651 &0.2905 &255.6154 &250.8157 &0.3124$_{\pm0.0029}$  \\
  \cmidrule{2-7}
 & \cellcolor{gray!20}$\text{NegBio-VAE}_\text{MC-G}$ &\cellcolor{gray!20}0.0180 &\cellcolor{gray!20}0.7132 & \cellcolor{gray!20}\textbf{127.5248} & \cellcolor{gray!20}\textbf{125.9497} \cellcolor{gray!20}
 &\cellcolor{gray!20}\textbf{0.1468$_{\pm0.0130}$}  \\
 & \cellcolor{gray!20}$\text{NegBio-VAE}_\text{MC-C}$ &\cellcolor{gray!20}0.0152 &\cellcolor{gray!20}0.7331 &\cellcolor{gray!20} 148.9795 &\cellcolor{gray!20}147.7799 &\cellcolor{gray!20}0.1688$_{\pm0.0128}$  \\
 & \cellcolor{gray!20}$\text{NegBio-VAE}_\text{DS-G}$ &\cellcolor{gray!20}0.0186 &\cellcolor{gray!20}0.6773 &\cellcolor{gray!20}\underline{133.0601} &\cellcolor{gray!20}\underline{132.8822} &\cellcolor{gray!20}\underline{0.1517}$_{\pm0.0149}$  \\
 & \cellcolor{gray!20}$\text{NegBio-VAE}_\text{DS-C}$ &\cellcolor{gray!20}\textbf{0.0144} &\cellcolor{gray!20}\textbf{0.7406} &\cellcolor{gray!20}155.5468 &\cellcolor{gray!20}154.1402 &\cellcolor{gray!20}0.1763$_{\pm0.0124}$ \\
 \midrule
\multirow{10}{*}{CIFAR$_{16\times 16}$}
 & $\mathcal{G}$-VAE   &0.1027 & 0.4495 & 72.0683 & 69.7067 & 0.0607$_{\pm0.0074}$  \\
 & $\mathcal{L}$-VAE  &0.0807 & 0.5079 & 91.1614 & 89.9475 & 0.0857$_{\pm0.0096}$   \\
 & $\mathcal{C}$-VAE  &0.0664 & 0.4755 &89.4235 &88.7412 & 0.0463$_{\pm0.0105}$  \\
 & $\mathcal{P}$-VAE   &0.0357 & \underline{0.6791} &60.3653 &59.1037 & 0.0582$_{\pm0.0098}$  \\
  % & $\text{NVAE}$ &\underline{0.0206} &0.6127 &315.1545 &314.5374 &0.3631$_{\pm0.0029}$ 
 % \\
  \cmidrule{2-7}
 & \cellcolor{gray!20}$\text{NegBio-VAE}_\text{MC-G}$ &\cellcolor{gray!20}0.0470 & \cellcolor{gray!20}0.6337 & \cellcolor{gray!20}\textbf{40.2788} & \cellcolor{gray!20}\textbf{39.8336} & \cellcolor{gray!20}\textbf{0.0348$_{\pm0.0065}$} \\
 & \cellcolor{gray!20}$\text{NegBio-VAE}_\text{MC-C}$ &\cellcolor{gray!20}0.0456 &\cellcolor{gray!20} 0.6429 &\cellcolor{gray!20} \cellcolor{gray!20}67.2898 & \cellcolor{gray!20}65.6569 &\cellcolor{gray!20}0.0727$_{\pm0.0096}$ \\
 & \cellcolor{gray!20}$\text{NegBio-VAE}_\text{DS-G}$ &\cellcolor{gray!20}0.0388 & \cellcolor{gray!20}0.6328 &\cellcolor{gray!20} \cellcolor{gray!20}\underline{41.7768} & \cellcolor{gray!20}\underline{41.1260} & \cellcolor{gray!20}\underline{0.0452}$_{\pm0.0080}$ \\
 & \cellcolor{gray!20}$\text{NegBio-VAE}_\text{DS-C}$ &\cellcolor{gray!20}\textbf{0.0189} & \cellcolor{gray!20}\textbf{0.8089} &\cellcolor{gray!20}64.9939 &\cellcolor{gray!20}63.6688 & \cellcolor{gray!20}0.0634$_{\pm0.0086}$  \\
\midrule
\multirow{10}{*}{CelebA$_{64\times64}$} 
 & $\mathcal{G}$-VAE   &0.4011 & 0.1772 &195.1377 &194.0974 &0.2758$_{\pm0.0192}$\\
 & $\mathcal{L}$-VAE  &0.3375 & 0.2161 &199.9303 &198.8191 &0.2655$_{\pm0.0117}$\\
 & $\mathcal{C}$-VAE  & 0.0774 & 0.4662 &166.2762 &165.7814 &0.1648$_{\pm0.0139}$ \\
 & $\mathcal{P}$-VAE   &0.0343 & \textbf{0.6354} &\underline{88.2312} & \underline{87.8107} & \textbf{0.0985}$_{\pm0.0088}$  \\
 % & $\text{NVAE}$ &\textbf{0.0141} &0.6323 &419.0647 &417.7907 &0.5742$_{\pm0.0011}$ \\
  \cmidrule{2-7}
 & \cellcolor{gray!20}$\text{NegBio-VAE}_\text{MC-G}$ &\cellcolor{gray!20}0.0451 &\cellcolor{gray!20} 0.5922 &\cellcolor{gray!20}89.7370 &\cellcolor{gray!20}88.4573 & \cellcolor{gray!20}0.1052$_{\pm0.0098}$ \\
 & \cellcolor{gray!20}$\text{NegBio-VAE}_\text{MC-C}$ &\cellcolor{gray!20}0.0373 &\cellcolor{gray!20}0.6165 &\cellcolor{gray!20}104.3009 &\cellcolor{gray!20}103.9739 & \cellcolor{gray!20}0.1165$_{\pm0.0084}$ \\
 & \cellcolor{gray!20}$\text{NegBio-VAE}_\text{DS-G}$ & \cellcolor{gray!20}0.0447& \cellcolor{gray!20}0.5982 &\cellcolor{gray!20}\textbf{84.2972} &\cellcolor{gray!20}\textbf{83.6357} & \cellcolor{gray!20}\underline{0.0992}$_{\pm0.0098}$ \\
 & \cellcolor{gray!20}$\text{NegBio-VAE}_\text{DS-C}$ & \cellcolor{gray!20}\underline{0.0341} 
 &\cellcolor{gray!20}\underline{0.6329} & \cellcolor{gray!20}92.8698& \cellcolor{gray!20}91.3648 & \cellcolor{gray!20}0.1069$_{\pm0.0098}$ \\
\bottomrule
\end{tabular}
}
\end{table*}

\section{Experiments}
\label{sec: exp}
% In this section, we compare NegBio-VAE with several VAE variants on four benchmark datasets, evaluating reconstruction quality, generative performance, and the expressiveness of latent representations for downstream tasks.

In this section, we compare NegBio-VAE with several well-known VAE variants on four standard benchmark datasets. These experiments are designed to evaluate the effectiveness of our model in terms of reconstruction quality, generative performance, and the expressiveness of the learned latent representations for downstream tasks. 

\begin{figure}[t]
\begin{center}
\begin{minipage}{0.3\linewidth}
\includegraphics[width=\linewidth]{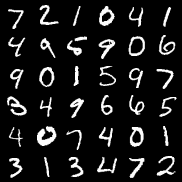}
\subcaption{Input images}\label{fig: mnist-initial-image}
\end{minipage}
\begin{minipage}{0.3\linewidth}
\includegraphics[width=\linewidth]{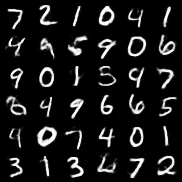}
\subcaption{MC-G}\label{fig: mnist-recon-mc+g}
\end{minipage}
\begin{minipage}{0.3\linewidth}
\includegraphics[width=\linewidth]{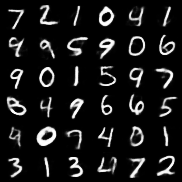}
\subcaption{MC-C}\label{fig: mnist-recon-mc+c}
\end{minipage}
\begin{minipage}{0.3\linewidth}
\includegraphics[width=\linewidth]{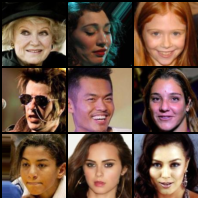}
\subcaption{Input images}\label{fig: celeba-initial-image}
\end{minipage}
\begin{minipage}{0.3\linewidth}
\includegraphics[width=\linewidth]{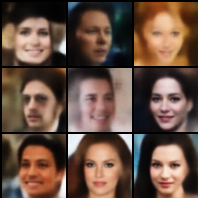}
\subcaption{MC-G}\label{fig: celeba-recon-mc+g}
\end{minipage}
\begin{minipage}{0.3\linewidth}
\includegraphics[width=\linewidth]{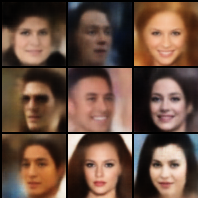}
\subcaption{MC-C}\label{fig: celeba-recon-mc+c}
\end{minipage}
\caption{Visual reconstruction results on the MNIST (top) and CelebA-64 (bottom) datasets using the MC-series variants of NegBio-VAE.}
\label{fig: maintex-recon-demo}
\end{center}
\vspace{-0.5cm}
\end{figure}

\subsection{Experimental Setup}
This section introduces the datasets, baselines, metrics, and implementation details.

\subsubsection{Datasets, Baselines and Metrics} 
We assess NegBio-VAE on four widely-used benchmark datasets: \textbf{MNIST}~\cite{lecun2010mnist,deng2012mnist}, \textbf{Fashion-MNIST}~\cite{fashion-mnist}, \textbf{CIFAR$_{16\times 16}$}~\cite{CIFAR10} and \textbf{CelebA-64}~\cite{celeba}. The model is compared with representative VAEs using either continuous or discrete latents. Continuous baselines include Gaussian VAE (\textbf{$\mathcal{G}$-VAE})~\cite{kingma2013auto}, Laplace VAE (\textbf{$\mathcal{L}$-VAE}), 
while discrete baselines include categorical VAE (\textbf{$\mathcal{C}$-VAE})~\cite{jang2017categorical} and Poisson VAE (\textbf{$\mathcal{P}$-VAE})~\cite{vafaii2024poisson}. We further examine four NegBio-VAE variants: \textbf{NegBio-VAE}$_\textbf{MC-G}$, \textbf{NegBio-VAE}$_\textbf{MC-C}$, \textbf{NegBio-VAE}$_\textbf{DS-G}$, and \textbf{NegBio-VAE}$_\textbf{DS-C}$, where \textbf{MC} denotes Monte Carlo, \textbf{DS} denotes dispersion sharing, and \textbf{G} and \textbf{C} indicate Gumbel-Softmax and continuous-time reparameterization. {It is worth noting that and we do not compare against certain strong baselines such as Nouveau VAE (\textbf{NVAE})~\cite{nave} and Very Deep VAE~\cite{child2021very} as these models are built upon hierarchical latent structures, making a direct comparison with our single-layer NegBio-VAE unfair. 
%The focus of this work is on demonstrating the advantages of NB-based discrete latent variables within the single-layer VAE setting.} Except for NVAE, which uses multi-layer latents, all baselines are single-layer, therefore, we use a single-layer NVAE for fair comparison. 
Model performance is evaluated from two perspectives: reconstruction and generation. For reconstruction, mean squared error (\textbf{MSE}) and structural similarity index (\textbf{SSIM}) measure fidelity and structural preservation after latent compression and decoding. For generation, Fréchet Inception Distance (\textbf{FID}) and Kernel Inception Distance (\textbf{KID}) quantify the discrepancy between generated and real data distributions, reflecting sample quality and diversity.

\subsubsection{Implementation.}
The encoder $\text{NB}(\mathbf{z};\, \mathbf{r} \odot \boldsymbol{\delta}_r(\mathbf{x}),\, \mathbf{p} \odot \boldsymbol{\delta}_p(\mathbf{x}))$ is implemented as a neural network that takes $\mathbf{x}$ as input and outputs $\boldsymbol{\delta}_p(\mathbf{x})$, optionally $\boldsymbol{\delta}_r(\mathbf{x})$. 
The decoder $p_\theta(\mathbf{x} \mid \mathbf{z})$ is modeled as a Gaussian distribution: $p_\theta(\mathbf{x} \mid \mathbf{z}) = \mathcal{N}\big(\mathbf{x}; f_\theta(\mathbf{z}), \sigma^2 \mathbf{I}\big)$, where $\sigma^2$ is a hyperparameter. This yields the reconstruction term: $\log p_\theta(\mathbf{x} \mid \mathbf{z}) = -\frac{1}{2\sigma^2} \Vert \mathbf{x} - f_\theta(\mathbf{z}) \Vert_2^2 + \text{const}$, which is equivalent to applying a coefficient $\beta = 2\sigma^2$ to the KL term in the ELBO, thereby balancing the trade-off between reconstruction and prior regularization. Unless otherwise specified, all VAEs use convolutional encoders and decoders, with the latent dimensionality fixed at 256.
%Experiments are implemented in PyTorch (Python 3.10.4) and run on two NVIDIA RTX 4090 GPUs (24GB each), and an Intel Xeon Gold 6430 CPU.

% \paragraph{Baselines.} 
% We compare NegBio-VAE with representative VAE models that employ either continuous or discrete latent variables. 
% The continuous baselines include the Gaussian VAE (\textbf{$\mathcal{G}$-VAE})~\cite{kingma2013auto}, Laplace VAE (\textbf{$\mathcal{L}$-VAE}), and Nouveau VAE (\textbf{NVAE})~\cite{nave}, 
% while the discrete baselines include the categorical VAE (\textbf{$\mathcal{C}$-VAE})~\cite{jang2017categorical} and Poisson VAE (\textbf{$\mathcal{P}$-VAE})~\cite{vafaii2024poisson}. 
% We further evaluate four variants of our model: \textbf{NegBio-VAE}$_\textbf{MC-G}$, \textbf{NegBio-VAE}$_\textbf{MC-C}$, \textbf{NegBio-VAE}$_\textbf{DS-G}$, and \textbf{NegBio-VAE}$_\textbf{DS-C}$. 
% Here, \textbf{MC} denotes Monte Carlo estimation, \textbf{DS} refers to dispersion sharing, and \textbf{G} and \textbf{C} represent Gumbel-Softmax and continuous-time reparameterization, respectively. 
% Notably, except for NVAE, which is a multi-layer latent variable VAE, all other baselines are single-layer latent variable VAEs. Therefore, to ensure a fair comparison, we use a single-layer version of NVAE in our experiments. 

\begin{table*}[t]
  \centering
  \caption{Evaluation of latent representations on MNIST for the fragmentation prediction task. Higher accuracy indicates more structured and generalizable latent representations. The best and second-best results are marked in \textbf{bold} and \underline{underlined}, respectively.}
  \label{tab: shattering dim}
   \resizebox{0.8\textwidth}{!}{
  \begin{tabular}{cccccc}
    \toprule
    Latent Dim           & Model             & \cellcolor{color_blue} Acc$\uparrow$(N=200) & \cellcolor{color_pink} Acc$\uparrow$(N=1000) & \cellcolor{color_yellow} Acc$\uparrow$(N=5000) & \cellcolor{blue!8}Acc$\uparrow$(Shat. Dim.) \\ \midrule
    \multirow{5}{*}{100} & $\mathcal{G}$-VAE & 0.790$_{\pm0.0070}$ & \textbf{0.914}$_{\pm0.0020}$ & \textbf{0.958}$_{\pm0.0020}$ & 0.890$_{\pm0.0050}$           \\
                         & $\mathcal{L}$-VAE & \underline{0.798}$_{\pm0.0090}$ & \underline{0.912}$_{\pm0.0020}$ & \textbf{0.958}$_{\pm0.0020}$ & \underline{0.892}$_{\pm0.0070}$      \\
                         & $\mathcal{C}$-VAE &0.783$_{\pm0.0070}$ & 0.896$_{\pm0.0030}$ & 0.941$_{\pm0.0040}$ & 0.886$_{\pm0.0070}$   \\
                         & $\mathcal{P}$-VAE & 0.736$_{\pm0.0110}$ & 0.888$_{\pm0.0020}$ & 0.947$_{\pm0.0030}$ & 0.862$_{\pm0.0070}$   \\
                         % & \text{NVAE} & & & & \\ 
                         \cmidrule(lr){2-6}
                         & \cellcolor{gray!20}NegBio-VAE        & \cellcolor{gray!20}\textbf{0.811$_{\pm0.0050}$} & \cellcolor{gray!20}\underline{0.912}$_{\pm0.0010}$ & \cellcolor{gray!20}\underline{0.955}$_{\pm0.0030}$ & \cellcolor{gray!20}\textbf{0.898$_{\pm0.0060}$} \\    \bottomrule
  \end{tabular}
  }
\end{table*}

\subsection{Reconstruction}
\label{sec: exp-performance-compare-reconstruction}
We first evaluate the reconstruction capability of the proposed method (\cref{tab: main result}). NegBio-VAE consistently achieves performance comparable to or better than existing {single-layer VAE baselines} across all datasets. Notably, the MC-C and DS-C variants attain the lowest MSE and highest SSIM on MNIST, Fashion-MNIST, and CIFAR$_{16\times16}$, demonstrating their ability to effectively preserve both structural information and fine-grained image details. On more complex datasets like CelebA-64, NegBio-VAE exhibits slightly higher reconstruction errors, likely due to the stronger regularization introduced by its biologically inspired priors. However, this also yields a more structured latent representation (\cref{sec: exp-performance-compare-latent}). Visual reconstruction results are presented in \cref{fig: maintex-recon-demo}, with additional results provided in Appendix C. 
% \cref{appendix: reconstruction image}. 

\begin{figure}[t]
\begin{center}
\begin{minipage}{0.3\linewidth}
\includegraphics[width=\linewidth]{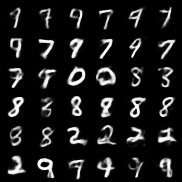}
\subcaption{Sample 1}\label{fig: mnist-generation-sample-1}
\end{minipage}
\begin{minipage}{0.3\linewidth}
\includegraphics[width=\linewidth]{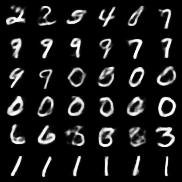}
\subcaption{Sample 2}\label{fig: mnist-generation-sample-2}
\end{minipage}
\begin{minipage}{0.3\linewidth}
\includegraphics[width=\linewidth]{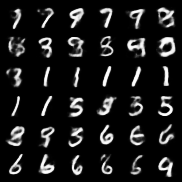}
\subcaption{Sample 3}\label{fig: mnist-generation-sample-3}
\end{minipage}
\begin{minipage}{0.3\linewidth}
\includegraphics[width=\linewidth]{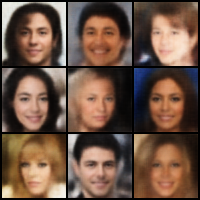}
\subcaption{Sample 1}\label{fig: celeba-generation-sample-1}
\end{minipage}
\begin{minipage}{0.3\linewidth}
\includegraphics[width=\linewidth]{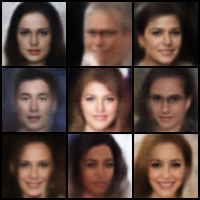}
\subcaption{Sample 2}\label{fig: celeba-generation-sample-2}
\end{minipage}
\begin{minipage}{0.3\linewidth}
\includegraphics[width=\linewidth]{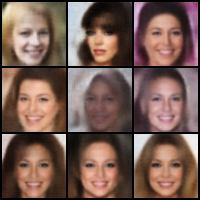}
\subcaption{Sample 3}\label{fig: celeba-generation-sample-3}
\end{minipage}
\caption{Samples randomly generated from the MNIST (top) and CelebA-64 (bottom) datasets using $\text{NegBio-VAE}_\text{MC-G}$.}
\label{fig: maintex-generation-demo}
\end{center}
\vspace{-0.5cm}
\end{figure}

\subsection{Generation}
\label{sec: exp-performance-compare-generation}
The generative performance of NegBio-VAE is assessed by sampling from the latent space. As shown in \cref{tab: main result}, NegBio-VAE significantly outperforms traditional VAEs, with consistently lower FID and KID scores. The MC-G variant shows the largest advantage, attaining the best results across nearly all datasets. For example, reducing FID to 39.8 on CIFAR$_{16\times16}$. These results indicate that the NB latent representation enhances flexibility, capturing richer and more diverse generative patterns. On CelebA-64, NegBio-VAE achieves the lowest FID and nearly the lowest KID compared to all baselines. Although NVAE is originally multi-layer, a single-layer version is used for fairness; even under this constraint, NegBio-VAE surpasses all baselines and could be extended to multi-layer latents to further improve performance. Visual results are shown in \cref{fig: maintex-generation-demo}, with additional results in Appendix D.

\begin{table*}[t]
  \centering
  \caption{Evaluation of latent representations on MNIST and CIFAR for the few-shot learning task. Higher accuracy indicates more structured and generalizable latent representations. The best and second-best results are marked in \textbf{bold} and \underline{underlined}, respectively.}
  \label{tab: few shot learning on MNIST}
  \resizebox{\textwidth}{!}{
  \begin{tabular}{cllllllll}
    \toprule
    \multirow{2}{*}{Model} & \multicolumn{4}{c}{Logistic Regression}                                                                                                             & \multicolumn{4}{c}{$k$NN} \\
    \cmidrule(lr){2-5} \cmidrule(lr){6-9}
                           & \multicolumn{1}{c}{Acc$\uparrow$(1-shot)} & \multicolumn{1}{c}{Acc$\uparrow$(5-shot)} & \multicolumn{1}{c}{Acc$\uparrow$(10-shot)} & \multicolumn{1}{c}{Acc$\uparrow$(20-shot)} & \multicolumn{1}{c}{Acc$\uparrow$(1-shot)} & \multicolumn{1}{c}{Acc$\uparrow$(5-shot)} & \multicolumn{1}{c}{Acc$\uparrow$(10-shot)} & \multicolumn{1}{c}{Acc$\uparrow$(20-shot)} \\ \midrule
    \rowcolor{red!40}\multicolumn{9}{c}{\textbf{{\textit{MNIST}}}}\\
    \midrule
    $\mathcal{G}$-VAE      & 0.409$_{\pm0.024}$              & 0.664$_{\pm0.022}$              & 0.736$_{\pm0.012}$               & 0.788$_{\pm0.010}$               & 0.228$_{\pm0.022}$              & 0.527$_{\pm0.015}$              & 0.653$_{\pm0.024}$               & 0.756$_{\pm0.008}$               \\
    $\mathcal{L}$-VAE      & 0.411$_{\pm0.024}$              & 0.666$_{\pm0.025}$              & 0.742$_{\pm0.012}$               & 0.794$_{\pm0.012}$               & 0.230$_{\pm0.036}$              & 0.534$_{\pm0.014}$              & 0.654$_{\pm0.026}$               & 0.760$_{\pm0.010}$               \\
    $\mathcal{C}$-VAE      & \underline{0.443}$_{\pm0.034}$              & 0.683$_{\pm0.030}$              & 0.755$_{\pm0.011}$               & 0.807$_{\pm0.012}$               & \textbf{0.283}$_{\pm0.032}$              & \textbf{0.593}$_{\pm0.018}$              & \textbf{0.714}$_{\pm0.013}$               & \textbf{0.791}$_{\pm0.011}$               \\
    $\mathcal{P}$-VAE      & 0.403$_{\pm0.031}$              & \underline{0.685}$_{\pm0.030}$              & \underline{0.760}$_{\pm0.015}$               & \underline{0.838}$_{\pm0.013}$               & 0.224$_{\pm0.023}$              & 0.498$_{\pm0.020}$              & 0.629$_{\pm0.010}$               & 0.720$_{\pm0.013}$               \\ 
    % \text{NVAE} &0.101$_{\pm0.003}$  &0.102$_{\pm0.003}$ &0.108$_{\pm0.004}$ &0.112$_{\pm0.004}$ &0.106$_{\pm0.005}$ &0.104$_{\pm0.003}$ &0.106$_{\pm0.005}$ &0.106$_{\pm0.004}$ \\
    \midrule
    \cellcolor{gray!20}NegBio-VAE             & \cellcolor{gray!20}\textbf{0.447$_{\pm0.031}$}     & \cellcolor{gray!20}\textbf{0.715$_{\pm0.027}$}     & \cellcolor{gray!20}\textbf{0.790$_{\pm0.011}$}      & \cellcolor{gray!20}\textbf{0.865$_{\pm0.011}$}      & \cellcolor{gray!20}\underline{0.273}$_{\pm0.020}$     & \cellcolor{gray!20}\underline{0.591}$_{\pm0.016}$     & \cellcolor{gray!20}\underline{0.710}$_{\pm0.011}$      & \cellcolor{gray!20}\underline{0.786}$_{\pm0.011}$      \\ \midrule
        \rowcolor{orange!40}\multicolumn{9}{c}{\textbf{{\textit{CIFAR}}}}\\
        \midrule
        $\mathcal{G}$-VAE      & 0.142$_{\pm0.013}$              & \underline{0.206}$_{\pm0.016}$              & 0.217$_{\pm0.014}$               & 0.238$_{\pm0.008}$               & 0.125$_{\pm0.015}$              & 0.144$_{\pm0.016}$              & 0.162$_{\pm0.011}$               & 0.182$_{\pm0.007}$               \\
    $\mathcal{L}$-VAE      & 0.138$_{\pm0.015}$              & 0.202$_{\pm0.016}$               & 0.213$_{\pm0.014}$               & 0.235$_{\pm0.007}$               & 0.124$_{\pm0.012}$              & 0.134$_{\pm0.014}$              & 0.151$_{\pm0.010}$               & 0.174$_{\pm0.007}$               \\
    $\mathcal{C}$-VAE      & \underline{0.158}$_{\pm0.025}$              & 0.190$_{\pm0.018}$              & 0.223$_{\pm0.011}$               & 0.240$_{\pm0.013}$               & \underline{0.131}$_{\pm0.018}$              & \underline{0.176}$_{\pm0.015}$              & \underline{0.194}$_{\pm0.010}$               & \underline{0.216}$_{\pm0.009}$               \\
    $\mathcal{P}$-VAE      & 0.154$_{\pm0.020}$              & 0.203$_{\pm0.016}$              & \underline{0.244}$_{\pm0.013}$               & \underline{0.261}$_{\pm0.012}$               & 0.120$_{\pm0.012}$              & 0.173$_{\pm0.015}$              & 0.188$_{\pm0.013}$               & 0.205$_{\pm0.010}$               \\ 
    % \text{NVAE} &0.142$_{\pm0.026}$ &.192$_{\pm0.014}$  &0.208$_{\pm0.011}$  &.216$_{\pm0.007}$ &0.139$_{\pm0.036}$  &0.152$_{\pm0.019}$ &0.164$_{\pm0.008}$ &0.170$_{\pm0.010}$ \\
    \midrule
    \cellcolor{gray!20}NegBio-VAE             & \cellcolor{gray!20}\textbf{0.167$_{\pm0.023}$}     & \cellcolor{gray!20}\textbf{0.221$_{\pm0.016}$}     & \cellcolor{gray!20}\textbf{0.255$_{\pm0.011}$}      & \cellcolor{gray!20}\textbf{0.266$_{\pm0.010}$}      & \cellcolor{gray!20}\textbf{0.133$_{\pm0.024}$}     & \cellcolor{gray!20}\textbf{0.192$_{\pm0.014}$}     & \cellcolor{gray!20}\textbf{0.207$_{\pm0.012}$}      & \cellcolor{gray!20}\textbf{0.233$_{\pm0.012}$}      \\ \bottomrule
    
  \end{tabular}
}
\end{table*}

\begin{figure*}[hbpt]
    \centering
    \begin{subfigure}{\linewidth}
        \centering
        \includegraphics[width=\linewidth]{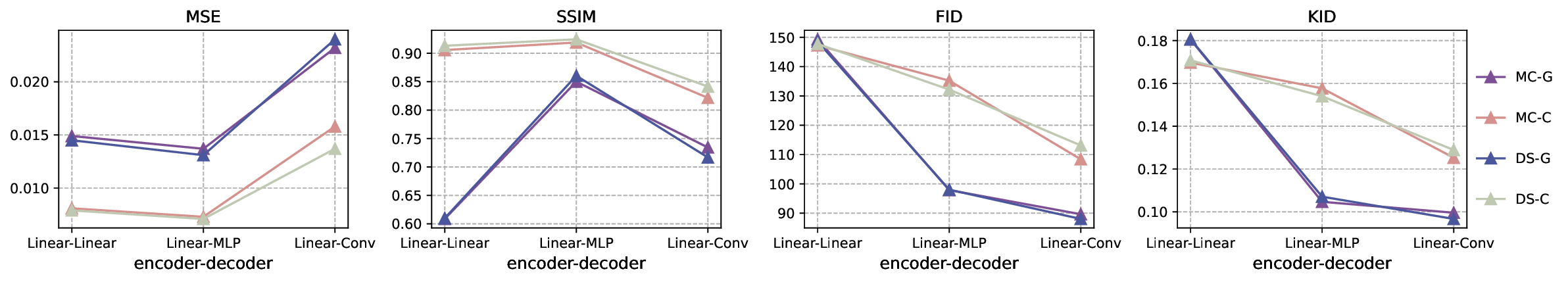}
        \caption{Ablation study of decoder architectures with encoder fixed as a linear network on reconstruction and generation quality.}
        \label{fig: ablation of encoder-decoder}
    \end{subfigure}

    \begin{subfigure}{\linewidth}
    \centering
    \includegraphics[width=\linewidth]{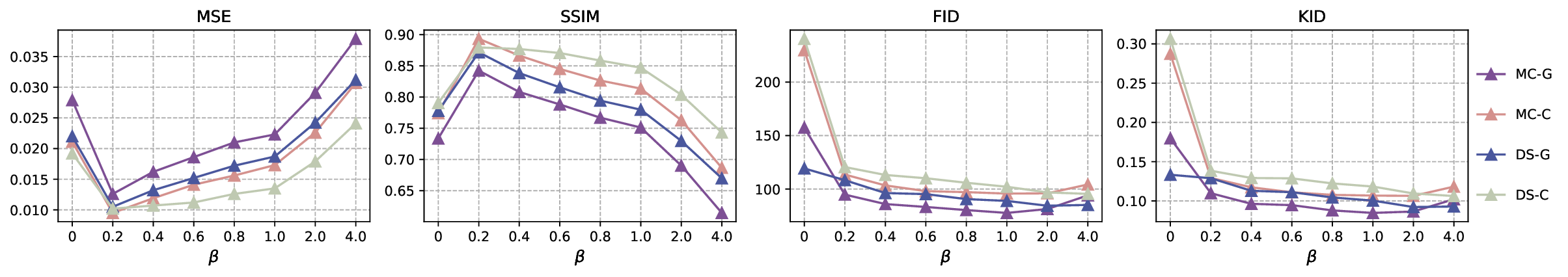}
    \caption{Ablation study of $\beta$ scaling on reconstruction and generation quality.}
     \label{fig: ablation of beta}
    \end{subfigure}

    \begin{subfigure}{\linewidth}
    \centering
    \includegraphics[width=\linewidth]{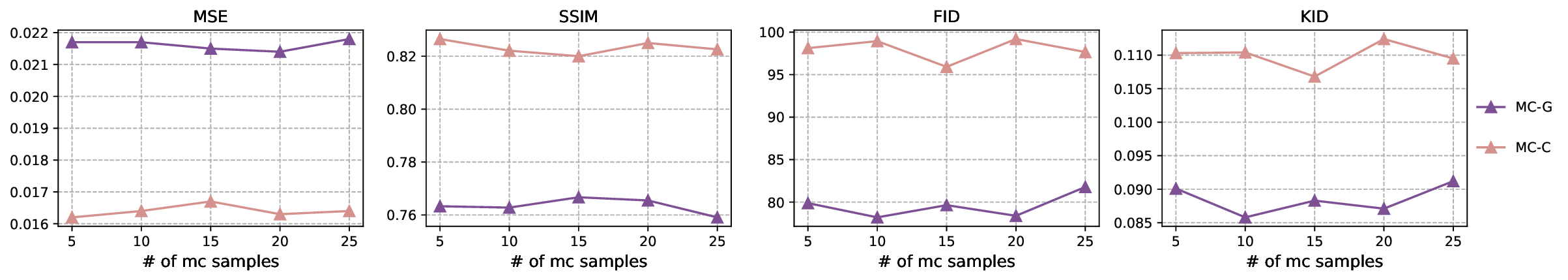}
    \caption{Ablation study on the number of Monte Carlo samples for reconstruction and generation quality.}
    \label{fig: ablation of mc sample}
    \end{subfigure}
    \caption{Ablation studies on decoder design, regularization strength, and the number of Monte Carlo samples in NegBio-VAE.}
\end{figure*}

\subsection{Latent Analysis}
\label{sec: exp-performance-compare-latent}
% To further evaluate the learned latent representations beyond reconstruction and generation, we assess their effectiveness in downstream tasks under two complementary settings: shattered experiments to test the robustness and discriminative power of the latent space under randomized labels, and few-shot experiments to evaluate how well the latents support classification with limited labeled samples. Together, these tasks provide a comprehensive measure of the model’s utility for downstream applications. Notably, we do not include NVAE in these comparisons because, even in its single-layer form, its latent variables differ fundamentally from traditional VAEs: rather than simple dense vectors, NVAE’s latents are spatially structured feature maps from a convolutional generative architecture, making them less compatible with standard downstream tasks such as classification or clustering.

We further evaluate latent representations on downstream tasks using two settings: fragmentation prediction, testing robustness under randomized labels, and few-shot learning, assessing classification with limited samples. {All experiments are repeated 10 times, and we report mean performance to support robust comparisons across models.} NVAE is excluded because even as a single-layer model, its latents are spatially structured feature maps rather than dense vectors, making them less compatible with standard tasks like classification or clustering.

\subsubsection{Fragmentation Prediction}
To evaluate robustness and discriminative power of the learned latent representations, we follow the setup in~\citet{vafaii2024poisson}, using MNIST with a fixed latent dimensionality of $100$ and convolutional encoder-decoders for all models. We randomly split the test set into two sets of 5,000 samples each and train logistic regression classifiers using $N = 200,\ 1000,\ 5000$ labeled samples from one set. We then report accuracy on the other set (\cref{tab: shattering dim}). For NegBio-VAE, we use the variant of DS-C. We also assess latent space structure via empirical shattering dimensionality~\citep{sd_bernardi2020geometry,sd_kaufman2022implications,sd_natrue_importance,vafaii2024poisson}, defined as the average binary classification accuracy over disjoint class partitions using linear classifiers. As shown in \cref{tab: shattering dim}, all models exhibit improved performance with increasing training samples. NegBio-VAE consistently ranks first or second, achieving 0.811 accuracy at $N=200$ and 0.898 under the shattering metric. This demonstrates that NB latents enhance separability and robustness even under severe label perturbations, whereas conventional models like $\mathcal{C}$-VAE are less resilient.
 
% achieves either the best or second-best scores across all settings, attaining an accuracy of $0.811$ at $N=200$ and $0.898$ under the shattering dimensionality metric.
% This demonstrates that the assumption of negative binomial latent variables effectively enhances the separability and robustness of the latent space, even under severe label perturbations. 
% In contrast, conventional variants such as $\mathcal{P}$-VAE and $\mathcal{C}$-VAE show weaker resistance to label shattering, suggesting that NegBio-VAE better preserves meaningful latent geometry against random supervision.

\subsubsection{Few-shot Learning}
We further evaluate the adaptability of the learned representations in low-data regimes through few-shot learning on MNIST and CIFAR$_{16\times 16}$. For each dataset, we use a $k$-shot setup with $k \in \{1, 5, 10, 20\}$, sampling $k$ labeled samples per class for training and evaluating on the test set. Two lightweight classifiers—logistic regression and $k$-nearest neighbors ($k$NN)—are used to assess the effectiveness of the latent representations. 
As shown in \cref{tab: few shot learning on MNIST}, NegBio-VAE consistently ranks first or second across all $k$ values. 
On MNIST, it achieves the highest accuracy with logistic regression, with the gap widening as labeled samples increase, e.g., 0.865 at 20-shot v.s. 0.838 for the best baseline. On CIFAR$_{16\times 16}$, NegBio-VAE similarly maintains superior performance, e.g., 0.266 at 20-shot v.s. 0.261 for the best baseline. These results show that NegBio-VAE learns more discriminative and transferable representations, enabling accurate classification with minimal supervision and robust transfer across datasets of varying complexity.

\subsection{Ablation Studies}
\label{sec: exp-performance-compare-ablation}
To further analyze NegBio-VAE, we perform ablation studies on MNIST, investigating the impact of the encoder–decoder design, $\beta$ scaling, and the number of Monte Carlo samples during training.

\subsubsection{Encoder-Decoder Architectures}
We compare encoder–decoder architectures using linear, multilayer perceptron (MLP), and convolutional networks. Results for linear encoders are shown in \cref{fig: ablation of encoder-decoder}, with further analyses in Appendix E. With a linear encoder, decoder choice strongly affects performance: MLP decoders achieve the lowest MSE and highest SSIM, convolutional decoders yield the best FID and KID, and linear decoders perform worst. These results highlight that enhancing decoder capacity, via nonlinear or convolutional architectures, is crucial for both reconstruction accuracy and generative quality.

\subsubsection{Effect of $\beta$ Scaling}
We investigate the effect of the scaling factor $\beta$ on NegBio-VAE performance (\cref{fig: ablation of beta}). Small $\beta$ values (0.2–0.4) yield the best reconstruction (lowest MSE, highest SSIM), especially for MC-C and DS-C variants. As $\beta$ increases, FID improves and peaks around $\beta$ = 1.0, indicating better generative fidelity. Beyond $\beta \ge 2.0$, reconstruction degrades and generative gains diminish. Overall, smaller $\beta$ favors reconstruction, larger $\beta$ favors generation, and intermediate values ($\approx$ 0.6–1.0) provide the best trade-off.

% When $\beta$ is small (e.g., 0.2–0.4), the models achieve the lowest reconstruction errors (MSE and SSIM), particularly under DS-C and MC-C settings, indicating that moderate regularization improves reconstruction quality. As $\beta$ increases, FID scores gradually improve, reaching their best values around $\beta$ = 1.0, which reflects enhanced generative fidelity. However, larger $\beta$ values ($\ge$ 2.0) lead to notable degradation in reconstruction while only marginally benefiting or even harming generative quality in certain settings. Overall, these results highlight a trade-off: smaller $\beta$ emphasizes reconstruction accuracy, whereas larger $\beta$ favors sample fidelity, with intermediate values ($\beta \approx$ 0.6–1.0) offering the most balanced performance.

\subsubsection{Effect of Number of MC Samples}
We examine the impact of the number of MC samples on model performance (\cref{fig: ablation of mc sample}). As the number of MC samples increases from 5 to 25, all metrics remain relatively stable for both MC-G and MC-C variants, indicating that the proposed model is robust to the sampling variance. Specifically, MC-C achieves lower MSE and higher SSIM (better reconstruction), while MC-G attains lower FID and KID (better generation). These results demonstrate that while increasing the number of MC samples provides only marginal gains, the model achieves a favorable balance between reconstruction and generation quality even with a few samples, confirming the effectiveness of our sampling strategy.

\section{Conclusions}
\label{sec: conclusion}
In this work, we presented NegBio-VAE, a generative model leveraging the NB distribution to capture overdispersed latent variables. By introducing a dispersion parameter, it extends beyond standard Poisson assumptions with minimal modification. Despite its simplicity, NegBio-VAE improves reconstruction and generation quality across benchmark datasets and outperforms existing VAE baselines in fidelity, generative quality, and the utility of latent representations for downstream tasks. While NegBio-VAE introduces greater flexibility in modeling overdispersed spike counts, the design choices, such as KL estimation and reparameterization, affect training and representations, a deeper theoretical understanding of these trade-offs remains open. Future work will explore adaptive reparameterization strategies based on data characteristics, and extend the framework to hierarchical latent structures similar to NVAE to further enhance model expressiveness.

\section*{Acknowledgments}
This work was supported by the NSFC Projects (Nos. 62506069, 62576346), the MOE Project of Key Research Institute of Humanities and Social Sciences (22JJD110001), the fundamental research funds for the central universities, and the research funds of Renmin University of China (24XNKJ13), and Beijing Advanced Innovation Center for Future Blockchain and Privacy Computing.

{
    \small
    \bibliographystyle{ieeenat_fullname}
    \bibliography{main}

@String(ICCV= {Int. Conf. Comput. Vis.})

@String(IJCAI = {IJCAI})

@String(AAAI = {AAAI})

@String(ICCV  = {ICCV})

@inproceedings{jang2017categorical,
  title={Categorical Reparameterization with Gumbel-Softmax},
  author={Jang, Eric and Gu, Shixiang and Poole, Ben},
  booktitle={International Conference on Learning Representations},
  year={2017}
}

@article{vafaii2024poisson,
  title={Poisson Variational Autoencoder},
  author={Vafaii, Hadi and Galor, Dekel and Yates, Jacob},
  journal={Advances in Neural Information Processing Systems},
  volume={37},
  pages={44871--44906},
  year={2024}
}

@article{tavanaei2019deep,
  title={Deep learning in spiking neural networks},
  author={Tavanaei, Amirhossein and Ghodrati, Masoud and Kheradpisheh, Saeed Reza and Masquelier, Timoth{\'e}e and Maida, Anthony},
  journal={Neural networks},
  volume={111},
  pages={47--63},
  year={2019},
  publisher={Elsevier}
}

@book{arbib2003handbook,
  title={The handbook of brain theory and neural networks},
  author={Arbib, Michael A},
  year={2003},
  publisher={MIT press}
}

@article{perkel1967neuronal,
  title={Neuronal spike trains and stochastic point processes: II. Simultaneous spike trains},
  author={Perkel, Donald H and Gerstein, George L and Moore, George P},
  journal={Biophysical journal},
  volume={7},
  number={4},
  pages={419--440},
  year={1967},
  publisher={Elsevier}
}

@article{mainen1995reliability,
  title={Reliability of spike timing in neocortical neurons},
  author={Mainen, Zachary F and Sejnowski, Terrence J},
  journal={Science},
  volume={268},
  number={5216},
  pages={1503--1506},
  year={1995},
  publisher={American Association for the Advancement of Science}
}

@article{bair1996temporal,
  title={Temporal precision of spike trains in extrastriate cortex of the behaving macaque monkey},
  author={Bair, Wyeth and Koch, Christof},
  journal={Neural computation},
  volume={8},
  number={6},
  pages={1185--1202},
  year={1996},
  publisher={MIT Press}
}

@article{gollisch2008rapid,
  title={Rapid neural coding in the retina with relative spike latencies},
  author={Gollisch, Tim and Meister, Markus},
  journal={science},
  volume={319},
  number={5866},
  pages={1108--1111},
  year={2008},
  publisher={American Association for the Advancement of Science}
}

@article{van2020brain,
  title={Brain-inspired replay for continual learning with artificial neural networks},
  author={Van de Ven, Gido M and Siegelmann, Hava T and Tolias, Andreas S},
  journal={Nature communications},
  volume={11},
  number={1},
  pages={4069},
  year={2020},
  publisher={Nature Publishing Group UK London}
}

@article{ghosh2009spiking,
  title={Spiking neural networks},
  author={Ghosh-Dastidar, Samanwoy and Adeli, Hojjat},
  journal={International journal of neural systems},
  volume={19},
  number={04},
  pages={295--308},
  year={2009},
  publisher={World Scientific}
}

@misc{kingma2013auto,
  title={Auto-encoding variational bayes},
  author={Kingma, Diederik P and Welling, Max and others},
  year={2013},
  publisher={Banff, Canada}
}

@article{marino2022predictive,
  title={Predictive coding, variational autoencoders, and biological connections},
  author={Marino, Joseph},
  journal={Neural Computation},
  volume={34},
  number={1},
  pages={1--44},
  year={2022},
  publisher={MIT Press One Rogers Street, Cambridge, MA 02142-1209, USA journals-info~…}
}

@article{vafaii2023hierarchical,
  title={Hierarchical VAEs provide a normative account of motion processing in the primate brain},
  author={Vafaii, Hadi and Yates, Jacob and Butts, Daniel},
  journal={Advances in Neural Information Processing Systems},
  volume={36},
  pages={46152--46190},
  year={2023}
}

@article{storrs2021unsupervised,
  title={Unsupervised learning predicts human perception and misperception of gloss},
  author={Storrs, Katherine R and Anderson, Barton L and Fleming, Roland W},
  journal={Nature Human Behaviour},
  volume={5},
  number={10},
  pages={1402--1417},
  year={2021},
  publisher={Nature Publishing Group UK London}
}

@article{taouali2016testing,
  title={Testing the odds of inherent vs. observed overdispersion in neural spike counts},
  author={Taouali, Wahiba and Benvenuti, Giacomo and Wallisch, Pascal and Chavane, Fr{\'e}d{\'e}ric and Perrinet, Laurent U},
  journal={Journal of neurophysiology},
  volume={115},
  number={1},
  pages={434--444},
  year={2016},
  publisher={American Physiological Society Bethesda, MD}
}

@article{moshitch2014using,
  title={Using Tweedie distributions for fitting spike count data},
  author={Moshitch, Dina and Nelken, Israel},
  journal={Journal of neuroscience methods},
  volume={225},
  pages={13--28},
  year={2014},
  publisher={Elsevier}
}

@article{stevenson2016flexible,
  title={Flexible models for spike count data with both over-and under-dispersion},
  author={Stevenson, Ian H},
  journal={Journal of computational neuroscience},
  volume={41},
  pages={29--43},
  year={2016},
  publisher={Springer}
}

@article{ross1985negative,
  title={The negative binomial distribution},
  author={Ross, GJS and Preece, DA},
  journal={Journal of the Royal Statistical Society: Series D (The Statistician)},
  volume={34},
  number={3},
  pages={323--335},
  year={1985},
  publisher={Wiley Online Library}
}

@inproceedings{cheng2020lisnn,
  title={LISNN: Improving spiking neural networks with lateral interactions for robust object recognition.},
  author={Cheng, Xiang and Hao, Yunzhe and Xu, Jiaming and Xu, Bo},
  booktitle={IJCAI},
  pages={1519--1525},
  year={2020},
  organization={Yokohama}
}

@inproceedings{zheng2021going,
  title={Going deeper with directly-trained larger spiking neural networks},
  author={Zheng, Hanle and Wu, Yujie and Deng, Lei and Hu, Yifan and Li, Guoqi},
  booktitle={Proceedings of the AAAI conference on artificial intelligence},
  volume={35},
  number={12},
  pages={11062--11070},
  year={2021}
}

@inproceedings{fang2021incorporating,
  title={Incorporating learnable membrane time constant to enhance learning of spiking neural networks},
  author={Fang, Wei and Yu, Zhaofei and Chen, Yanqi and Masquelier, Timoth{\'e}e and Huang, Tiejun and Tian, Yonghong},
  booktitle={Proceedings of the IEEE/CVF international conference on computer vision},
  pages={2661--2671},
  year={2021}
}

@article{li2021differentiable,
  title={Differentiable spike: Rethinking gradient-descent for training spiking neural networks},
  author={Li, Yuhang and Guo, Yufei and Zhang, Shanghang and Deng, Shikuang and Hai, Yongqing and Gu, Shi},
  journal={Advances in neural information processing systems},
  volume={34},
  pages={23426--23439},
  year={2021}
}

@inproceedings{kamata2022fully,
  title={Fully spiking variational autoencoder},
  author={Kamata, Hiromichi and Mukuta, Yusuke and Harada, Tatsuya},
  booktitle={Proceedings of the AAAI conference on artificial intelligence},
  volume={36},
  number={6},
  pages={7059--7067},
  year={2022}
}

@inproceedings{yadav2025differentially,
  title={Differentially Private Spiking Variational Autoencoder},
  author={Yadav, Srishti and Pundhir, Anshul and Goyal, Tanish and Raman, Balasubramanian and Kumar, Sanjeev},
  booktitle={International Conference on Pattern Recognition},
  pages={96--112},
  year={2025},
  organization={Springer}
}

@inproceedings{liu2024spiking,
  title={Spiking-diffusion: Vector quantized discrete diffusion model with spiking neural networks},
  author={Liu, Mingxuan and Gan, Jie and Wen, Rui and Li, Tao and Chen, Yongli and Chen, Hong},
  booktitle={2024 5th International Conference on Computer, Big Data and Artificial Intelligence (ICCBD+ AI)},
  pages={627--631},
  year={2024},
  organization={IEEE}
}

@inproceedings{kotariya2022spiking,
  title={Spiking-GAN: A spiking generative adversarial network using time-to-first-spike coding},
  author={Kotariya, Vineet and Ganguly, Udayan},
  booktitle={2022 International Joint Conference on Neural Networks (IJCNN)},
  pages={1--7},
  year={2022},
  organization={IEEE}
}

@article{rosenfeld2022spiking,
  title={Spiking generative adversarial networks with a neural network discriminator: Local training, bayesian models, and continual meta-learning},
  author={Rosenfeld, Bleema and Simeone, Osvaldo and Rajendran, Bipin},
  journal={IEEE Transactions on Computers},
  volume={71},
  number={11},
  pages={2778--2791},
  year={2022},
  publisher={IEEE}
}

@article{feng2024spiking,
  title={Spiking generative adversarial network with attention scoring decoding},
  author={Feng, Linghao and Zhao, Dongcheng and Zeng, Yi},
  journal={Neural Networks},
  volume={178},
  pages={106423},
  year={2024},
  publisher={Elsevier}
}

@inproceedings{cao2024spiking,
  title={Spiking denoising diffusion probabilistic models},
  author={Cao, Jiahang and Wang, Ziqing and Guo, Hanzhong and Cheng, Hao and Zhang, Qiang and Xu, Renjing},
  booktitle={Proceedings of the IEEE/CVF winter conference on applications of computer vision},
  pages={4912--4921},
  year={2024}
}

@article{kapoor2024latent,
  title={Latent diffusion for neural spiking data},
  author={Kapoor, Jaivardhan and Schulz, Auguste and Vetter, Julius and Pei, Felix and Gao, Richard and Macke, Jakob H},
  journal={Advances in Neural Information Processing Systems},
  volume={37},
  pages={118119--118154},
  year={2024}
}

@article{dupont2018learning,
  title={Learning disentangled joint continuous and discrete representations},
  author={Dupont, Emilien},
  journal={Advances in neural information processing systems},
  volume={31},
  year={2018}
}

@article{zhan2023esvae,
  title={Esvae: An efficient spiking variational autoencoder with reparameterizable poisson spiking sampling},
  author={Zhan, Qiugang and Tao, Ran and Xie, Xiurui and Liu, Guisong and Zhang, Malu and Tang, Huajin and Yang, Yang},
  journal={arXiv preprint arXiv:2310.14839},
  year={2023}
}

@article{van2017neural,
  title={Neural discrete representation learning},
  author={Van Den Oord, Aaron and Vinyals, Oriol and others},
  journal={Advances in neural information processing systems},
  volume={30},
  year={2017}
}

@inproceedings{fortuinsom,
  title={SOM-VAE: Interpretable Discrete Representation Learning on Time Series},
  author={Fortuin, Vincent and H{\"u}ser, Matthias and Locatello, Francesco and Strathmann, Heiko and R{\"a}tsch, Gunnar},
  booktitle={International Conference on Learning Representations},
  year={2019}
}

@inproceedings{rolfe2017discrete,
  title={Discrete Variational Autoencoders},
  author={Rolfe, Jason Tyler},
  booktitle={International Conference on Learning Representations},
  year={2017}
}

@inproceedings{zhao2020variational,
  title={Variational autoencoders for sparse and overdispersed discrete data},
  author={Zhao, He and Rai, Piyush and Du, Lan and Buntine, Wray and Phung, Dinh and Zhou, Mingyuan},
  booktitle={International conference on artificial intelligence and statistics},
  pages={1684--1694},
  year={2020},
  organization={PMLR}
}

@inproceedings{polykovskiy2020deterministic,
  title={Deterministic decoding for discrete data in variational autoencoders},
  author={Polykovskiy, Daniil and Vetrov, Dmitry},
  booktitle={International conference on artificial intelligence and statistics},
  pages={3046--3056},
  year={2020},
  organization={PMLR}
}

@article{pillow2012fully,
  title={Fully Bayesian inference for neural models with negative-binomial spiking},
  author={Pillow, Jonathan and Scott, James},
  journal={Advances in neural information processing systems},
  volume={25},
  year={2012}
}

@article{di2011nbp,
  title={The NBP negative binomial model for assessing differential gene expression from RNA-Seq},
  author={Di, Yanming and Schafer, Daniel W and Cumbie, Jason S and Chang, Jeff H},
  journal={Statistical applications in genetics and molecular biology},
  volume={10},
  number={1},
  year={2011},
  publisher={De Gruyter}
}

@article{zhou2013negative,
  title={Negative binomial process count and mixture modeling},
  author={Zhou, Mingyuan and Carin, Lawrence},
  journal={IEEE Transactions on Pattern Analysis and Machine Intelligence},
  volume={37},
  number={2},
  pages={307--320},
  year={2013},
  publisher={IEEE}
}

@article{fashion-mnist,
  title={Fashion-mnist: a novel image dataset for benchmarking machine learning algorithms},
  author={Xiao, Han and Rasul, Kashif and Vollgraf, Roland},
  journal={arXiv preprint arXiv:1708.07747},
  year={2017}
}

@INPROCEEDINGS{celeba,
  author={Liu, Ziwei and Luo, Ping and Wang, Xiaogang and Tang, Xiaoou},
  booktitle={2015 IEEE International Conference on Computer Vision (ICCV)}, 
  title={Deep Learning Face Attributes in the Wild}, 
  year={2015},
  volume={},
  number={},
  pages={3730-3738},
  doi={10.1109/ICCV.2015.425}}

@misc{kingma2014autoencoding,
      title={Auto-Encoding Variational Bayes}, 
      author={Diederik P Kingma and Max Welling},
      year={2014},
      eprint={1312.6114},
      archivePrefix={arXiv},
      primaryClass={stat.ML}
}

@TECHREPORT{CIFAR10,
    author = {Alex Krizhevsky},
    title = {Learning multiple layers of features from tiny images},
    year = {2009}
}

@article{sd_natrue_importance,
author={Rigotti, Mattia and Barak, Omri and Warden, Melissa R. and Wang, Xiao-Jing and Daw, Nathaniel D. and Miller, Earl K. and Fusi, Stefano},
year = {2013},
month = {05},
pages = {585-590},
title = {The importance of mixed selectivity in complex cognitive tasks},
volume = {497},
journal = {Nature},
doi = {https://doi.org/10.1038/nature12160}
}

@article{sd_bernardi2020geometry,
  title={The geometry of abstraction in the hippocampus and prefrontal cortex},
  author={Bernardi, Silvia and Benna, Marcus K. and Rigotti, Mattia and Munuera, J{\'e}r{\^o}me and Fusi, Stefano and Salzman, C. Daniel},
  journal={Cell},
  volume={183},
  number={4},
  pages={954--967.e21},
  year={2020},
  month={Nov},
  doi={10.1016/j.cell.2020.09.031},
  pmid={33058757},
  pmcid={PMC8451959},
  publisher={Cell Press}
}

@article{sd_kaufman2022implications,
  title={The implications of categorical and category-free mixed selectivity on representational geometries},
  author={Kaufman, Matthew T. and Benna, Marcus K. and Rigotti, Mattia and Stefanini, Fabio and Fusi, Stefano and Churchland, Anne K.},
  journal={Current Opinion in Neurobiology},
  volume={77},
  pages={102644},
  year={2022},
  month={Dec},
  doi={10.1016/j.conb.2022.102644},
  pmid={36332415},
  publisher={Elsevier}
}

@article{deng2012mnist,
  title={The mnist database of handwritten digit images for machine learning research},
  author={Deng, Li},
  journal={IEEE Signal Processing Magazine},
  volume={29},
  number={6},
  pages={141--142},
  year={2012},
  publisher={IEEE}
}

@article{lecun2010mnist,
  title={MNIST handwritten digit database},
  author={LeCun, Yann and Cortes, Corinna and Burges, CJ},
  journal={ATT Labs [Online]. Available: http://yann.lecun.com/exdb/mnist},
  volume={2},
  year={2010}
}

@inproceedings{nave,
 author = {Vahdat, Arash and Kautz, Jan},
 booktitle = {Advances in Neural Information Processing Systems},
 editor = {H. Larochelle and M. Ranzato and R. Hadsell and M.F. Balcan and H. Lin},
 pages = {19667--19679},
 publisher = {Curran Associates, Inc.},
 title = {NVAE: A Deep Hierarchical Variational Autoencoder},
 url = {https://proceedings.neurips.cc/paper_files/paper/2020/file/e3b21256183cf7c2c7a66be163579d37-Paper.pdf},
 volume = {33},
 year = {2020}
}

@article{shadlen1998variable,
  title={The Variable Discharge of Cortical Neurons: Implications for Connectivity, Computation, and Information Coding},
  author={Shadlen, Michael N. and Newsome, William T.},
  journal={Journal of Neuroscience},
  volume={18},
  number={10},
  pages={3870--3896},
  year={1998},
  publisher={Society for Neuroscience},
  doi={10.1523/JNEUROSCI.18-10-03870.1998}
}

@article{berry1997structure,
  title={The structure and precision of retinal spike trains},
  author={Berry, Michael J. and Warland, David K. and Meister, Markus},
  journal={Proceedings of the National Academy of Sciences},
  volume={94},
  number={10},
  pages={5411--5416},
  year={1997},
  publisher={National Academy of Sciences},
  doi={10.1073/pnas.94.10.5411}
}

@article{goris2014partitioning,
  title={Partitioning neuronal variability},
  author={Goris, Robbe L. T. and Movshon, J. Anthony and Simoncelli, Eero P.},
  journal={Nature Neuroscience},
  volume={17},
  pages={858--865},
  year={2014},
  publisher={Nature Publishing Group},
  doi={10.1038/nn.3711}
}

@inproceedings{
child2021very,
title={Very Deep {\{}VAE{\}}s Generalize Autoregressive Models and Can Outperform Them on Images},
author={Rewon Child},
booktitle={International Conference on Learning Representations},
year={2021},
url={https://openreview.net/forum?id=RLRXCV6DbEJ}
}

@inproceedings{NEURIPS2018_92c8c96e,
 author = {Figurnov, Mikhail and Mohamed, Shakir and Mnih, Andriy},
 booktitle = {Advances in Neural Information Processing Systems},
 editor = {S. Bengio and H. Wallach and H. Larochelle and K. Grauman and N. Cesa-Bianchi and R. Garnett},
 pages = {},
 publisher = {Curran Associates, Inc.},
 title = {Implicit Reparameterization Gradients},
 url = {https://proceedings.neurips.cc/paper_files/paper/2018/file/92c8c96e4c37100777c7190b76d28233-Paper.pdf},
 volume = {31},
 year = {2018}
}
}

\appendix
\clearpage
\appendix
\setcounter{page}{1}
\maketitlesupplementary
\section{Derivation of KL Term}
\label{sec: appendix-kl-term}
We derive an analytical expression for the KL divergence between two negative binomial distributions under the assumption that the encoder does not modify the dispersion parameter (i.e., $\delta_r = 1$). the univariate negative binomial distribution, given dispersion $r$ and success probability $p$, is defined as:
\begin{equation*}
    \text{NB}(z;r,p) = \begin{pmatrix} z+r-1 \\ z \end{pmatrix}(1-p)^zp^r.
\end{equation*}
Substituting this into the KL divergence for a single $z$ yields:
\begin{equation*}
    \begin{aligned}
        &\mathcal{D}_{\text{KL}}(q||p) = \mathbb{E}_{z\sim q}[\log \frac{q}{p}]\\
        & = \mathbb{E}_{z\sim q}\left [\log \frac{\begin{pmatrix}
        z+r\delta_r-1 \\ 
        z
    \end{pmatrix}(1-p\delta_p)^{z}(p\delta_p)^{r\delta_r}}{\begin{pmatrix}
        z+r-1 \\ 
        z
    \end{pmatrix}(1-p)^zp^r}\right ]\\
    %     &= \mathbb{E}_{z\sim q}\left[\log \left [\begin{pmatrix}
    %     z+r\delta_r-1 \\ 
    %     z
    % \end{pmatrix}\right ] - \log \left [\begin{pmatrix}
    %     z+r-1 \\ 
    %     z
    % \end{pmatrix}\right ] 
    % + (r\delta_r) \log [p\delta_p] - r \log p + z \log \left( \frac{1-p\delta_p}{1-p}\right)
    % \right ]\\
    & = \mathbb{E}_{z\sim q} \left[ r \log \frac{p \delta_p}{p} + z \log \left(\frac{1 - p\delta_p}{1 - p}\right)
    \right]    \\
    & = r \log \frac{p \delta_p}{p} + \mathbb{E}_{z\sim q} \left[ z \log \left(\frac{1 - p\delta_p}{1 - p}\right)
    \right] \\
    & = r \log \delta_p + \log \left(\frac{1 - p\delta_p}{1 - p}\right)\mathbb{E}_{z\sim q} [z]  \\
    & = r \log \delta_p + r \frac{1 - p\delta_p}{p\delta_p} \log \left(\frac{1 - p\delta_p}{1 - p}\right) \\
    & = r\left[ \log \delta_p + \frac{1-p\delta_p}{p\delta_p} \log \left(\frac{1 - p\delta_p}{1 - p}\right)
        \right]\\
    & = rg(p,\delta_p).
    \end{aligned}
\end{equation*}

Taking the logarithm of the terms involve binomial coefficients and computing the expectation with respect to the posterior makes the KL divergence intractable. As a result, Monte Carlo sampling or variational approximation techniques are typically required, which often introduce high variance in the gradient estimates or rely on additional approximating assumptions and can lead to unstable or biased training. To make the expression tractable, we introduce a simplifying assumption: $ \delta_r = 1$, i.e., the encoder does not adjust the prior parameter $r$, and thus the posterior and prior share the same dispersion parameter. This assumption is reasonable because the NB distribution is parameterized by both $r$ and $p$. Therefore, even when $r$ is fixed, we can still adjust the distribution (i.e., its mean and variance) by varying $p$. This leads to a closed-form approximation:
\begin{equation*}
\mathcal{D}_{\text{KL}}(q || p) = r \left [ \log \delta_p + \frac{1 - p\delta_p}{p\delta_p} \log \left( \frac{1 - p\delta_p}{1 - p} \right) \right ],
\end{equation*}
which we denote as $r  g(p, \delta_p)$, where:
\begin{gather*}
    g(a,b):= \log b + \frac{1-ab}{ab} \log \left[\frac{1-ab}{1-a}\right], \\ a\in (0,1), \quad b>0.
\end{gather*}
This expression is simple, interpretable and has useful boundary properties. When $\delta_p = 1$ (i.e., the encoder does not shift $p$), $g(a, 1) = 0$, and the KL divergence vanishes. As $ab \to 0$ (i.e., posterior sparsity increases), the KL grows rapidly, penalizing excessive deviation from the prior. This behavior mirrors that of $\mathcal{P}$-VAE, which strongly discourages low-rate posterior collapse.
% \begin{figure}[h]
%     \centering
%     \includegraphics[width=0.5\linewidth]{plots/g_function.pdf}
%     \caption{Plot of $g(a, b)$ with respect to $b$, for fixed $a$.}
%     \label{fig:g_function}
% \end{figure}
While $\mathcal{P}$-VAE already provides an elegant analysis of sparsity through its KL structure, we do not emphasize this aspect in the main text. However, our formulation shares the same desirable sparsity behavior: when $\delta_p$ approaches 0, the KL diverges, discouraging extreme posterior sparsification. Moreover, our formulation retains an analytical form even for overdispersed distributions, enabling tractable training without Monte Carlo estimation. 

Similar to the $\mathcal{P}$-VAE~\cite{vafaii2024poisson}, which analyzes the behavior of its KL divergence near the prior via a Taylor expansion of the function $f(\delta_r) = 1 - \delta_r + \delta_r \log \delta_r$, we perform a similar analysis for the closed-form KL term in NegBio-VAE. To better understand the behavior of the closed-form KL divergence near the prior, we expand $g(a, b)$ at $b = 1 + \epsilon$, with $\epsilon \ll 1$:
\begin{equation}
    g(a, 1 + \epsilon) \approx \frac{a}{2(1 - a)} \epsilon^2 + \mathcal{O}(\epsilon^3).
\end{equation}
Thus, when $\delta_p = 1 + \epsilon$, the KL becomes:
\begin{equation*}
    \mathcal{D}_{\text{KL}} \approx r \cdot \frac{a}{2(1 - a)} \epsilon^2.
\end{equation*}
This reveals that, like in $\mathcal{P}$-VAE, the KL divergence grows quadratically near the prior, encouraging smooth and stable optimization. However, our formulation provides a tunable growth rate via the parameter $a = p$, allowing more flexible control over sparsity regularization. Unlike Poisson VAEs, which assume equal mean and variance, our negative binomial model accommodates overdispersion and remains analytically tractable—enabling stable training without Monte Carlo approximation. These properties make our approach better suited for modeling realistic, variable spike-based neural activity.

\section{Implementation Details}
\label{}
We include all the implementation details in this section, including the sampling techniques and detailed experimental settings.
\subsection{Sampling Techniques}

\label{sec: appendix-implementation-sampling}
We adopt two sampling techniques for our model, while we have introduced the main idea in the main text, for completeness, we include the details here:

(1) \textbf{Gumbel-Softmax Relaxation} 
This method approximates discrete Poisson sampling using continuous relaxation. 
\begin{enumerate}
    \item 
    % Truncated Poisson Distribution: 
    Limit the maximum count value to \( Z_{\text{max}} \). 
    \item 
    % Log Probability Calculation: 
    Compute the log-probability for \( z = 0, 1, \ldots, Z_{\text{max}} \), 
    \[
        \log \text{Poi}(z) = z \log \lambda - \lambda - \log\Gamma(z + 1).
    \]
    \item 
    % Add Gumbel Noise: 
    For each \( z \), generate noise \( \epsilon_z \sim \text{Gumbel}(0, 1) \). 
    \item 
    % Generate Soft Samples: 
    Apply the Gumbel-Softmax trick with temperature \( \tau \), 
    \[
        \tilde{z} = \sum_{z=0}^{Z_{\text{max}}} z \cdot \mathrm{softmax}\left( \frac{\log \text{Poi}(z) + \epsilon_z}{\tau} \right), 
    \]
    where \( \tau \to 0 \) recovers discrete sampling. 
\end{enumerate}
The proof of this reparameterization can be found in \citet{jang2017categorical}, and will not be repeated here. 

(2) \textbf{Continuous-Time Simulation} 
% \citep{vafaii2024poisson} 
This method models Poisson processes with intensity $\lambda$ on $[0,1]$ using exponentially distributed inter-arrival times. 
\begin{enumerate}
    \item Sample inter-arrival times from an exponential distribution: 
    \[
        \{s_i\}_{i=1}^M \sim \text{Exponential}(\lambda),
    \]
    where $M$ is a sufficiently large integer, the exponential distribution is easily reparameterized and PyTorch contains an implementation. 
    
    \item Accumulate inter-arrival times: 
    \[
        S_n = \sum_{i=1}^n s_i, \quad 1\leq n \leq M. 
    \]
    \item Soft count of events: 
    \[
        \tilde{z} = \sum_{n=1}^M\sigma\left(\frac{1-S_n}{\tau}\right), 
    \]
    where \( \tau \to 0 \) recovers discrete sampling. 
\end{enumerate}
This reparameterization exploits the relationship between the Poisson distribution and the Poisson process. We can generate Poisson counts from \( \text{Poi}(\lambda) \) by counting events on a homogeneous Poisson process with intensity \( \lambda \) over the interval \( [0,1] \).

To verify that both proposed reparameterization methods can successfully generate valid count samples from the NB distribution, we generate 1000 samples from each method using $r=20$, $p=0.5$, and temperature $\tau=0.1$, and the empirical distribution is shown in \cref{fig:reparam_sampling}. Both methods produce plausible count distributions with unimodal structure and similar mean values, confirming that each method can successfully approximate NB samples in a differentiable manner. This validates their use as practical reparameterization techniques for NegBio-VAE.

\begin{figure}
    \centering
    \includegraphics[width=0.8\linewidth]{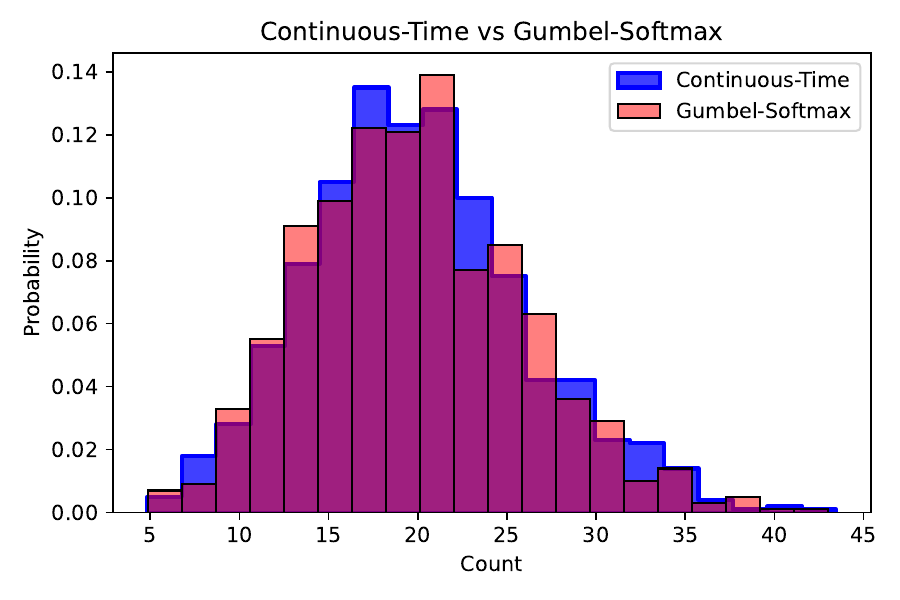}
    \caption{Empirical distributions of negative binomial samples generated using Continuous-Time Simulation and Gumbel-Softmax relaxation. Both methods successfully approximate count-valued outputs consistent with NB sampling behavior, validating their use as differentiable reparameterization strategies. Each method generates 1000 samples using parameters $r = 20$, $p = 0.5$, and temperature $\tau = 0.1$.}
    \label{fig:reparam_sampling}
    \vspace{-0.2cm}
\end{figure}

\subsection{Experimental Implementation}
In this section, we provide additional implementation details that complement the experiments described in the main text.
\subsubsection{Datasets}
We implement all data loading pipelines using PyTorch Lightning’s LightningDataModule interface, ensuring consistent structure across datasets. For each dataset, we apply preprocessing transformations tailored to its modality and input requirements. All preprocessing logic is encapsulated in a shared function 
\texttt{get\_transform()}, which dynamically composes transformations based on dataset type, grayscale conversion, data flattening, and augmentation flags.
\begin{itemize}
    \item MNIST: Grayscale images are normalized to $[0,1]$, using \texttt{transforms.ToTensor()} without additional augmentation. Images are optionally flattened if required by the encoder structure.
    \item Fashion-MNIST: Following the same preprocessing as MNIST, grayscale images are normalized to the $[0,1]$ range using \texttt{transforms.ToTensor()} without any additional augmentation. Images are optionally flattened when required by the encoder architecture.

    \item CIFAR$_{16\times 16}$: We use CIFAR-10 as a base dataset and uniformly downsample all images to 16 $\times$ 16 resolution. Color images are normalized to $[-1,1]$ using mean and standard deviation $(0.5,0.5,0.5)$ and are optionally augmented via horizontal flipping. 
    \item CelebA-64: For CelebA-64, RGB face images are resized to $64\times64$ and normalized to the $[-1,1]$ range using \texttt{transforms.ToTensor()} followed by \texttt{transforms.Resize(64)}. To ensure consistency across samples, no additional augmentation is applied. When required by the encoder architecture, images are optionally flattened or converted to latent representations.
\end{itemize}

\subsubsection{Encoder and Decoder Architectures}
Our model supports interchangeable encoder and decoder architectures to accommodate various data modalities and representation structures. Following the same setting in ~\cite{vafaii2024poisson}, we include linear, MLP, and convolutional-based designs. 
\begin{itemize}
    \item Linear Encoder: A single fully connected layer that maps flattened inputs to the latent space. Optionally applies weight normalization.
    \item MLP Encoder: A two-layer perceptron with an intermediate residual dense layer followed by a liner projection. This encoder supports flexible nonlinearity and is used when richer transformations are required from vectorized inputs.
    \item  Convolutional Encoder: A two-stage convolutional network with ReLU activations, followed by flattening and a fully connected projection. It adapts the input channel size (1 for graysacle datasets, 3 for RGB), and auto-computes the flattening shape based on the dataset resolution. LayerNorm is optionally applied to the final latent layer.
    \item Linear Decoder: Mirrors the linear encoder with a fully connected layer projecting latent vectors to pixel space. Output is passed through either a Sigmoid or Tanh nonlinearity depending on the expected pixel scale.
    \item MLP Decoder: A three-layer feedforward network with residual blocks and configurable nonlinearity (e.g., Swish), designed for richer reconstructions from compact latent codes. The final layer uses Sigmoid or Tanh.
    \item Convolutional Decoder: Used in image-based settings, this decoder first expands latent vectors through a fully connected layer into a low-resolution feature map, then applies transposed convolutions to upscale to the desired image size. The initial size is determined by dataset type (e.g., $7\times7$ for MNIST, $4\times4$ for CIFAR$_{16\times 16}$ ).
\end{itemize}
\subsubsection{Shattering Dimensionality}
To quantitatively evaluate the geometry of the learned latent space, we compute the shattering dimensionality following prior work. Specifically, we measure how well the latent space supports linear separation across all balanced binary label partitions. Concretely, given a label set such as digits 0-9, we enumerate all disjoint splits into two non-overlapping and balanced class groups, where each partition defines a binary classification task. For each split, we relabel samples in one group as class 0 and those in the other group as class 1, producing binary-labeled data. For a dataset with 10 classes (e.g., MNIST), we generate all possible balanced, disjoint 5-vs-5 class splits. This results in a total of 252 unique binary classification tasks, corresponding to all combinations of 5 classes out of 10 without regard to class order or labeling symmetry. A linear classifier (e.g., logistic regression) is then trained on latent representations from a subset of the training data. The classification accuracy is computed on the validation set, and the final shattering dimensionality is taken as the average accuracy across all 252 tasks. The implementation uses the \texttt{itertools.combinations} function to enumerate all unique 5-class subsets, and constructs their complementary partitions to define the 5-vs-5 classification groups.

\begin{figure*}[htbp]
    \begin{subfigure}{\linewidth}
        \centering
        \includegraphics[width=\linewidth]{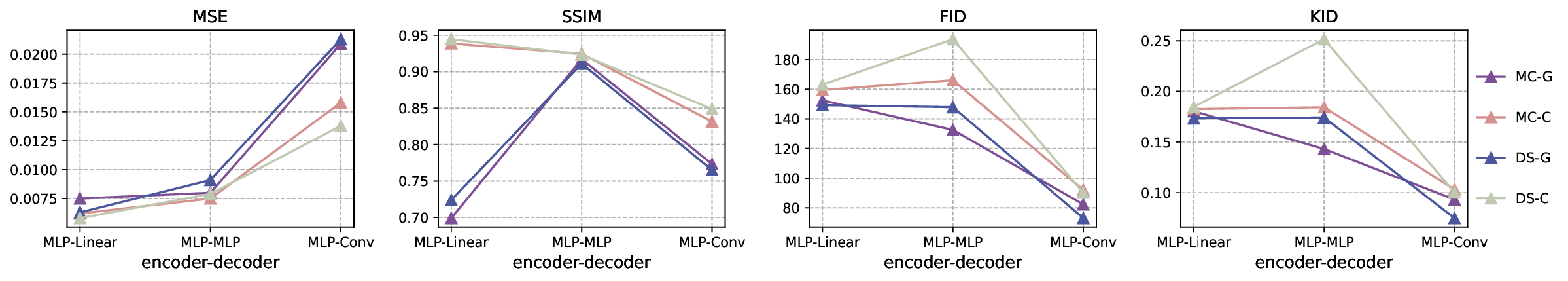}
        \caption{MLP encoder}
    \end{subfigure}

    \begin{subfigure}{\linewidth}
        \centering
        \includegraphics[width=\linewidth]{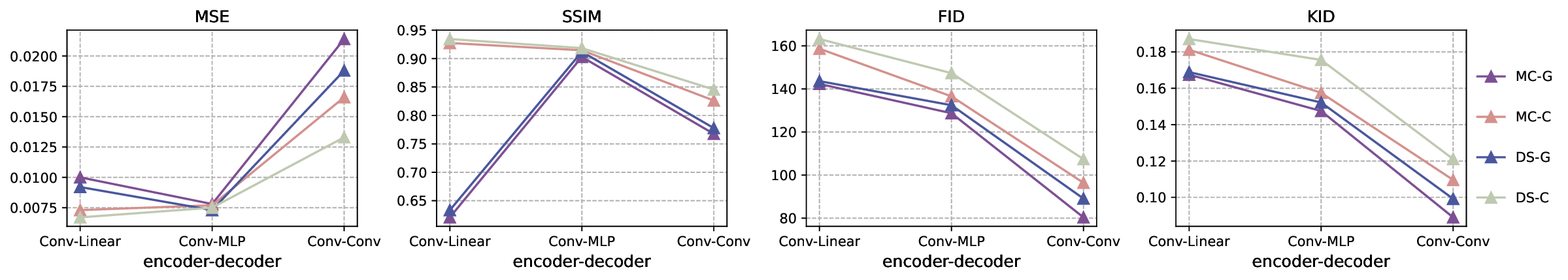}
        \caption{Convolutional encoder}
    \end{subfigure}
    \caption{Ablation study of encoder-decoder architectures on MNIST with four variants of NegBio-VAE.}
    \label{fig: appendix ablation of encoder-decoder}
    \vspace{-0.1cm}
\end{figure*}

\begin{figure*}[htbp]
    \centering
    \includegraphics[width=\linewidth]{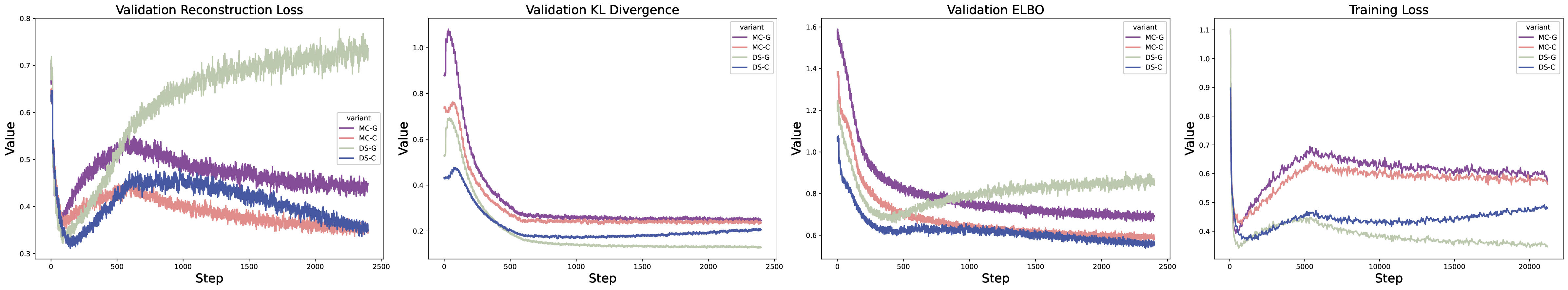}
    \caption{Comparison of four NegBio-VAE variants (MC-G, MC-C, DS-C, DS-G) across validation reconstruction loss, validation KL divergence, validation ELBO and training loss.}
    \label{fig:loss_compare}
    \vspace{-0.1cm}
\end{figure*}

\subsubsection{SSIM Computation Details}
To further assess the perceptual quality of reconstructed images, we employ the structural similarity index (SSIM) as a complementary metric. SSIM evaluates image similarity by considering changes in luminance, contrast, and structural information between two images. Given two images $x$ and $y$, SSIM is defined as:
\begin{equation*}
    \mathrm{SSIM}(x,y) = \frac{(2\mu_x\mu_y+C_1)(2\sigma _{xy}+C_2)}{(\mu^2_x+\mu^2_y+C_1)(\sigma^2_x+\sigma ^2_y+C_2)}, 
\end{equation*}
where $\mu_x$ and $\mu_y$ denote the mean intensities, $\sigma_x^2$ and $\sigma_y^2$ are the variances, and $\sigma_{xy}$ represents the covariance between $x$ and $y$. Constants $C_1$ and $C_2$ are used to stabilize the division. A higher SSIM indicated greater perceptual similarity between the reconstructed and ground-truth images. 
\subsubsection{FID Computation Details}
To quantitatively evaluate the visual fidelity and distributional similarity of generated images, we adopt the 
Fréchet Inception Distance (FID) as a standard evaluation metric. FID measures the distance between the real and generated image distributions in the feature space of a pretrained Inception network. Specifically, it assumes both distributions are Gaussian, and computes the Fréchet distance between them as:
\begin{equation*}
    \mathrm{FID} = \|\mu_{\text{real}} - \mu_{\text{gen}}\|^2 
+ \mathrm{Tr}\left( 
\Sigma_{\text{real}} + \Sigma_{\text{gen}} 
- 2(\Sigma_{\text{real}} \Sigma_{\text{gen}})^{1/2} 
\right),
\end{equation*}
where $\mu_{\text{real}}$, $\Sigma_{\text{real}}$ and $\mu_{\text{gen}}$, $\Sigma_{\text{gen}}$ denote the mean and covariance of the real and generated feature activations, respectively. A lower FID score indicates that the generated samples are more similar to the real data in terms of both image quality and diversity.

\subsubsection{KID Computation Details}
We also report the Kernel Inception Distance (KID) to evaluate the distributional alignment between real and generated images. Unlike FID, KID computes the squared Maximum Mean Discrepancy (MMD) between Inception representations of real and generated samples using a polynomial kernel. Formally, it is defined as:
\begin{equation*}
    \mathrm{KID}=\left \|\frac{1}{n_r}\sum_{i=1}^{n_r}\phi (x_i) - \frac{1}{n_g}\sum_{j=1}^{n_g}\phi(y_j)      \right \|  ^2,
\end{equation*}
where $\phi(\cdot)$ denotes the feature embedding from a pretrained Inception network, and $n_r$, $n_g$ are the numbers of real and generated samples, respectively. A smaller KID score indicated that the generated distribution is closer to the real one. Compare with FID, KID is unbiased and more reliable when computed with limited sample sizes.

\section{Visualization of Image Reconstruction Results}
\label{appendix: reconstruction image}
To further evaluate the reconstruction performance of NegBio-VAE, we present reconstructed samples from multiple datasets in \cref{app.mnist}, \cref{app.fmnist}, \cref{app.cifar}, and \cref{app.celeba}. As observed, our method accurately preserves fine-grained structural details across different datasets. For example, it retains the subtle gaps surrounding digits in the MNIST dataset and effectively reconstructs the contours and texture details of clothing in the Fashion-MNIST dataset. These observations demonstrate the strong capability of NegBio-VAE in modeling discrete structural information.

\begin{table*}[t]
  \centering
  \caption{Evaluation of latent representations on MNIST. Higher accuracy and shattering dimensionality indicate more structured and generalizable latent.}
  \label{tab: appendix shattering dim}
   \resizebox{0.8\textwidth}{!}{
  \begin{tabular}{cccccc}
    \toprule
    Latent Dim           & Model             & \cellcolor{color_blue} Acc$\uparrow$(N=200) & \cellcolor{color_pink} Acc$\uparrow$(N=1000) & \cellcolor{color_yellow} Acc$\uparrow$(N=5000) & \cellcolor{blue!8} Acc$\uparrow$(Shat. Dim.) \\ \midrule
    \multirow{5}{*}{10}  & $\mathcal{G}$-VAE & 0.726$_{\pm0.0015}$     & 0.798$_{\pm0.0020}$      & 0.844$_{\pm0.0040}$      & 0.851$_{\pm0.0050}$             \\
                         & $\mathcal{L}$-VAE &0.647$_{\pm0.0160}$   &0.733$_{\pm0.0080}$   & 0.781$_{\pm0.0040}$   &0.811$_{\pm0.0070}$             \\
                         & $\mathcal{C}$-VAE & 0.728$_{\pm0.0190}$  & 0.812$_{\pm0.0060}$  & 0.855$_{\pm0.0020}$   &0.856$_{\pm0.0090}$               \\
                         & $\mathcal{P}$-VAE & \underline{0.747}$_{\pm0.0180}$   & \textbf{0.836$_{\pm0.0030}$}  & \textbf{0.883$_{\pm0.0040}$}  & \textbf{0.865$_{\pm0.0080}$}             \\ \cmidrule(lr){2-6}
                         & \cellcolor{gray!20}NegBio-VAE        & \cellcolor{gray!20}\textbf{0.749$_{\pm0.0150}$}      & \cellcolor{gray!20}\underline{0.830}$_{\pm0.0010}$       & \cellcolor{gray!20}\underline{0.878}$_{\pm0.0020}$       & \cellcolor{gray!20}\underline{0.862}$_{\pm0.0080}$     
                         \\ \midrule
    \multirow{5}{*}{50}  & $\mathcal{G}$-VAE & 0.819$_{\pm0.0090}$ & \textbf{0.922}$_{\pm0.0030}$ & \textbf{0.960}$_{\pm0.0020}$ & \underline{0.903}$_{\pm0.0070}$   
         \\
                         & $\mathcal{L}$-VAE & \underline{0.822}$_{\pm0.0080}$ & \underline{0.921}$_{\pm0.0020}$ & \textbf{0.960}$_{\pm0.0030}$ & \underline{0.903}$_{\pm0.0060}$ 
        \\
                         & $\mathcal{C}$-VAE & 0.784$_{\pm0.0090}$ & 0.888$_{\pm0.0040}$ & 0.936$_{\pm0.0030}$ & 0.887$_{\pm0.0060}$ 
       \\
                         & $\mathcal{P}$-VAE & 0.760$_{\pm0.0130}$ & 0.897$_{\pm0.0030}$ & 0.951$_{\pm0.0020}$ & 0.872$_{\pm0.0060}$ 
   \\ \cmidrule(lr){2-6}
                         & \cellcolor{gray!20}NegBio-VAE         & \cellcolor{gray!20}\textbf{0.826$_{\pm0.0070}$} & \cellcolor{gray!20}0.914$_{\pm0.0020}$ & \cellcolor{gray!20}\underline{0.952}$_{\pm0.0030}$ & \cellcolor{gray!20}\textbf{0.904$_{\pm0.0070}$}     \\ \midrule
    \multirow{5}{*}{100} & $\mathcal{G}$-VAE & 0.790$_{\pm0.0070}$ & \textbf{0.914}$_{\pm0.0020}$ & \textbf{0.958}$_{\pm0.0020}$ & 0.890$_{\pm0.0050}$           \\
                         & $\mathcal{L}$-VAE & \underline{0.798}$_{\pm0.0090}$ & \underline{0.912}$_{\pm0.0020}$ & \textbf{0.958}$_{\pm0.0020}$ & \underline{0.892}$_{\pm0.0070}$      \\
                         & $\mathcal{C}$-VAE &0.783$_{\pm0.0070}$ & 0.896$_{\pm0.0030}$ & 0.941$_{\pm0.0040}$ & 0.886$_{\pm0.0070}$   \\
                         & $\mathcal{P}$-VAE & 0.736$_{\pm0.0110}$ & 0.888$_{\pm0.0020}$ & 0.947$_{\pm0.0030}$ & 0.862$_{\pm0.0070}$   \\
                         & \cellcolor{gray!20}NegBio-VAE        & \cellcolor{gray!20}\textbf{0.811$_{\pm0.0050}$} & \cellcolor{gray!20}\underline{0.912}$_{\pm0.0010}$ & \cellcolor{gray!20}\underline{0.955}$_{\pm0.0030}$ & \cellcolor{gray!20}\textbf{0.898$_{\pm0.0060}$} \\    \bottomrule
  \end{tabular}
  }
\end{table*}

\section{Visualization of Image Generation Results}
\label{appendix: generation image}
We further provide qualitative image generation results in \cref{app.mnist-sample}, \cref{app.fmnist-sample}, \cref{app.cifar-sample}, and \cref{app.celeba-sample}.
The generated samples demonstrate that NegBio-VAE produces diverse and visually coherent outputs, effectively capturing meaningful variations within the data distribution. These results further confirm the strong generative ability of the model and the well-structured organization of its learned latent space.

\section{Additional Experiments}
This section presents additional experimental results, including the latent representation analysis, further evaluations on NegBio-VAE architectures variants, and a detailed analysis of the loss evolution during training.

\subsection{Additional Results on Latent Analysis}
\label{appendix: additional res on latent analysis}
\cref{tab: appendix shattering dim} extends the shattering test results to different latent dimensions ($10$, $50$, and $100$) on MNIST. Across all configurations, NegBio-VAE consistently achieves the best performance, particularly under limited data ($N=200$), demonstrating its superior sample efficiency and robustness. Notably, NegBio-VAE attains the highest shattering dimensionalities ($0.862$, $0.904$, and $0.898$ for latent dimensions $10$, $50$, and $100$, respectively), indicating a more structured and stable latent space under randomized supervision. As the latent dimension increases, all models exhibit performance gains; however, NegBio-VAE maintains smoother improvement trends, suggesting stronger regularization and more biologically consistent representation learning.

\subsection{Additional Results on VAE Architecture Variants}
\label{appendix: ablation of encoder-decoder}
\cref{fig: appendix ablation of encoder-decoder} presents an ablation study on different encoder-decoder architectures using the MNIST dataset. In this study, the latent dimension of all variants is fixed at 256, and both MLP and convolutional architectures are used as encoders. Experimental results show that for MC-based methods, the MLP encoder generally achieves the lowest reconstruction error (MSE), while the convolutional architecture achieves higher generation quality (i.e., lower FID). Notably, the combination of an MLP encoder and a convolutional decoder achieves the best balance between reconstruction and generation, outperforming purely linear or convolutional designs. Similar trends are observed for DS-based methods: the MLP encoder consistently achieves the lowest MSE, while the convolutional decoder achieves competitive or even superior FID scores.

\subsection{Loss Dynamics Across NegBio-VAE Variants}
\label{sec: appendix-loss-dynamics}
\cref{fig:loss_compare} presents a comparison of the training dynamics for four NegBio-VAE variants across different loss terms (validation reconstruction loss, validation KL, validation ELBO and train loss). We observe notable differences in both the convergence rate and the smoothness of the trajectories, which reflect the influence of the KL estimation method and the reparametrization strategy. Overall, models with MC KL estimation (MC-G and MC-C) exhibit higher variance in the loss curves, with visible oscillations due to the stochastic nature of the Monte Carlo method (which introduces noise into the gradient updates as we have discussed in \cref{sec:method-kl-divergence}). In contrast, DS-based variants (DS-C and DS-G), which leverage closed-form KL computation via dispersion sharing, show smoother and more stable curves, suggesting better optimization stability. Comparing reparameterization strategies, models using continuous-time simulation (C) (DS-C and MC-C) tend to achieve lower reconstruction loss and faster ELBO convergence than their Gumbel-softmax (G). This suggests that the continuous-time approach offers a more expressive and stable mechanism for modeling spike-like latent representations. In particular, DS-C demonstrates the most stable and efficient convergence across all loss types, with consistently smooth trajectories and lower final values.

% \begin{figure*}[t]
%   \centering
%   \includegraphics[width=\linewidth]{plots/recons/mnisthightlight.pdf}
%   \caption{Reconstructions from different VAEs (256 dims) on MNIST. Despite using the same latent dimensionality, NegBio-VAE recovers sharper digit structures (e.g., the gap in \textbf{0} and the loop in \textbf{4}) that others miss. This advantage stems from its ability to model overdispersed latent spike counts, providing greater variance flexibility and enabling sharper distinctions in high-variability regions.}
%   \label{fig:mnist-highlight}
% \end{figure*}

\begin{figure*}[t]
\begin{center}
\begin{minipage}{0.196\textwidth}
\includegraphics[width=\linewidth]{plots/recons/mnist/initial.png}
\subcaption{Input images}\label{fig: mnist-initial}
\end{minipage}
\begin{minipage}{0.196\textwidth}
\includegraphics[width=\linewidth]{plots/recons/mnist/mc+g.png}
\subcaption{MC-G}\label{fig: mnist-mc+g}
\end{minipage}
\begin{minipage}{0.196\textwidth}
\includegraphics[width=\linewidth]{plots/recons/mnist/mc+c.png}
\subcaption{MC-C}\label{fig: mnist-mc+c}
\end{minipage}
\begin{minipage}{0.196\textwidth}
\includegraphics[width=\linewidth]{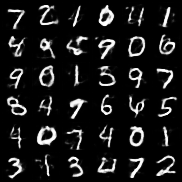}
\subcaption{DS-G}\label{fig: mnist-ds+g}
\end{minipage}
\begin{minipage}{0.196\textwidth}
\includegraphics[width=\linewidth]{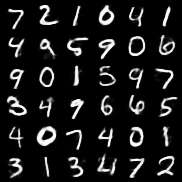}
\subcaption{DS-C}\label{fig: mnist-ds+c}
\end{minipage}
\caption{Reconstruction results on the MNIST dataset. The leftmost column shows the original images, while the remaining columns display the reconstructed images generated by NegBio-VAE.}
\label{app.mnist}
\end{center}
\end{figure*}

\begin{figure*}[t]
\begin{center}
\begin{minipage}{0.196\textwidth}
\includegraphics[width=\linewidth]{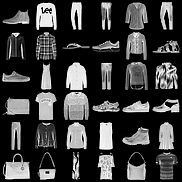}
\subcaption{Input images}\label{fig: fmnist-initial}
\end{minipage}
\begin{minipage}{0.196\textwidth}
\includegraphics[width=\linewidth]{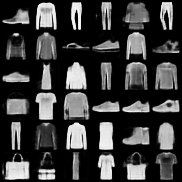}
\subcaption{MC-G}\label{fig: fmnist-mc+g}
\end{minipage}
\begin{minipage}{0.196\textwidth}
\includegraphics[width=\linewidth]{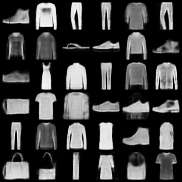}
\subcaption{MC-C}\label{fig: fmnist-mc+c}
\end{minipage}
\begin{minipage}{0.196\textwidth}
\includegraphics[width=\linewidth]{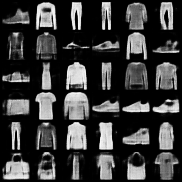}
\subcaption{DS-G}\label{fig: fmnist-ds+g}
\end{minipage}
\begin{minipage}{0.196\textwidth}
\includegraphics[width=\linewidth]{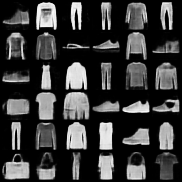}
\subcaption{DS-C}\label{fig: fmnist-ds+c}
\end{minipage}
\caption{Reconstruction results on the Fashion-MNIST dataset. The leftmost column shows the original images, while the remaining columns display the reconstructed images generated by NegBio-VAE.}
\label{app.fmnist}
\end{center}
\end{figure*}

\begin{figure*}[t]
\begin{center}
\begin{minipage}{0.196\textwidth}
\includegraphics[width=\linewidth]{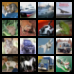}
\subcaption{Input images}\label{fig: cifar-initial}
\end{minipage}
\begin{minipage}{0.196\textwidth}
\includegraphics[width=\linewidth]{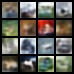}
\subcaption{MC-G}\label{fig: cifar-mc+g}
\end{minipage}
\begin{minipage}{0.196\textwidth}
\includegraphics[width=\linewidth]{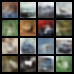}
\subcaption{MC-C}\label{fig: cifar-mc+c}
\end{minipage}
\begin{minipage}{0.196\textwidth}
\includegraphics[width=\linewidth]{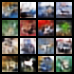}
\subcaption{DS-G}\label{fig: cifar-ds+g}
\end{minipage}
\begin{minipage}{0.196\textwidth}
\includegraphics[width=\linewidth]{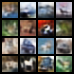}
\subcaption{DS-C}\label{fig: cifar-ds+c}
\end{minipage}
\caption{Reconstruction results on the CIFAR$_{16\times16}$ dataset. The leftmost column shows the original images, while the remaining columns display the reconstructed images generated by NegBio-VAE.} 
\label{app.cifar}
\end{center}
\end{figure*}

\begin{figure*}[t]
\begin{center}
\begin{minipage}{0.196\textwidth}
\includegraphics[width=\linewidth]{plots/recons/celeba64/initial.png}
\subcaption{Input images}\label{fig: celeba-initial}
\end{minipage}
\begin{minipage}{0.196\textwidth}
\includegraphics[width=\linewidth]{plots/recons/celeba64/mc-g.png}
\subcaption{MC-G}\label{fig: celeba-mc+g}
\end{minipage}
\begin{minipage}{0.196\textwidth}
\includegraphics[width=\linewidth]{plots/recons/celeba64/mc-c.png}
\subcaption{MC-C}\label{fig: celeba-mc+c}
\end{minipage}
\begin{minipage}{0.196\textwidth}
\includegraphics[width=\linewidth]{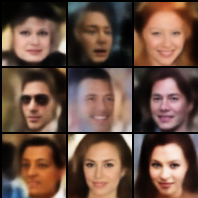}
\subcaption{DS-G}\label{fig: celeba-ds+g}
\end{minipage}
\begin{minipage}{0.196\textwidth}
\includegraphics[width=\linewidth]{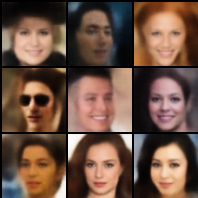}
\subcaption{DS-C}\label{fig: celeba-ds+c}
\end{minipage}
\caption{Reconstruction results on the CelebA-64 dataset. The leftmost column shows the original images, while the remaining columns display the reconstructed images generated by NegBio-VAE.} 
\label{app.celeba}
\end{center}
\end{figure*}

\begin{figure*}[t]
\begin{center}
\begin{minipage}{0.2\textwidth}
\includegraphics[width=\linewidth]{plots/generation/mnist/sample-1.png}
\subcaption{Sample 1}\label{fig: app-mnist-sample-1}
\end{minipage}
\begin{minipage}{0.2\textwidth}
\includegraphics[width=\linewidth]{plots/generation/mnist/sample-2.png}
\subcaption{Sample 2}\label{fig: app-mnist-sample-2}
\end{minipage}
\begin{minipage}{0.2\textwidth}
\includegraphics[width=\linewidth]{plots/generation/mnist/sample-3.png}
\subcaption{Sample 3}\label{fig: app-mnist-sample-3}
\end{minipage}
\begin{minipage}{0.2\textwidth}
\includegraphics[width=\linewidth]{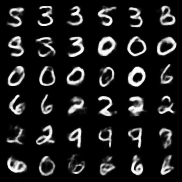}
\subcaption{Sample 4}\label{fig: app-mnist-sample-4}
\end{minipage}
\caption{Randomly generated samples on the MNIST dataset using NegBio-VAE. Each image is generated from a different random latent variable 
$z$ under identical model settings, illustrating the NegBio-VAE’s ability to produce diverse and realistic samples.}
\label{app.mnist-sample}
\end{center}
\end{figure*}

\begin{figure*}[t]
\begin{center}
\begin{minipage}{0.2\textwidth}
\includegraphics[width=\linewidth]{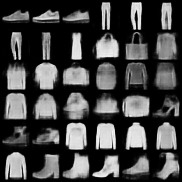}
\subcaption{Sample 1}\label{fig: fmnist-sample-1}
\end{minipage}
\begin{minipage}{0.2\textwidth}
\includegraphics[width=\linewidth]{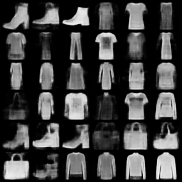}
\subcaption{Sample 2}\label{fig: fmnist-sample-2}
\end{minipage}
\begin{minipage}{0.2\textwidth}
\includegraphics[width=\linewidth]{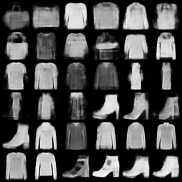}
\subcaption{Sample 3}\label{fig: fmnist-sample-3}
\end{minipage}
\begin{minipage}{0.2\textwidth}
\includegraphics[width=\linewidth]{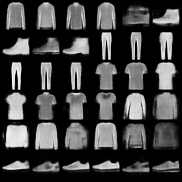}
\subcaption{Sample 4}\label{fig: fmnist-sample-4}
\end{minipage}
\caption{Randomly generated samples on the Fashion-MNIST dataset using NegBio-VAE. Each image is generated from a different random latent variable 
$z$ under identical model settings, illustrating the NegBio-VAE’s ability to produce diverse and realistic samples.}
\label{app.fmnist-sample}
\end{center}
\end{figure*}

\begin{figure*}[t]
\begin{center}
\begin{minipage}{0.2\textwidth}
\includegraphics[width=\linewidth]{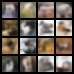}
\subcaption{Sample 1}\label{fig: cifar-sample-1}
\end{minipage}
\begin{minipage}{0.2\textwidth}
\includegraphics[width=\linewidth]{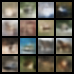}
\subcaption{Sample 2}\label{fig: cifar-sample-2}
\end{minipage}
\begin{minipage}{0.2\textwidth}
\includegraphics[width=\linewidth]{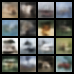}
\subcaption{Sample 3}\label{fig: cifar-sample-3}
\end{minipage}
\begin{minipage}{0.2\textwidth}
\includegraphics[width=\linewidth]{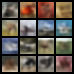}
\subcaption{Sample 4}\label{fig: cifar-sample-4}
\end{minipage}
\caption{Randomly generated samples on the CIFAR$_{16\times16}$ dataset using NegBio-VAE. Each image is generated from a different random latent variable 
$z$ under identical model settings, illustrating the NegBio-VAE’s ability to produce diverse and realistic samples.}
\label{app.cifar-sample}
\end{center}
\end{figure*}

\begin{figure*}[t]
\begin{center}
\begin{minipage}{0.2\textwidth}
\includegraphics[width=\linewidth]{plots/generation/celeba/sample-1.png}
\subcaption{Sample 1}\label{fig: celeba-sample-1}
\end{minipage}
\begin{minipage}{0.2\textwidth}
\includegraphics[width=\linewidth]{plots/generation/celeba/sample-2.png}
\subcaption{Sample 2}\label{fig: celeba-sample-2}
\end{minipage}
\begin{minipage}{0.2\textwidth}
\includegraphics[width=\linewidth]{plots/generation/celeba/sample-3.png}
\subcaption{Sample 3}\label{fig: celeba-sample-3}
\end{minipage}
\begin{minipage}{0.2\textwidth}
\includegraphics[width=\linewidth]{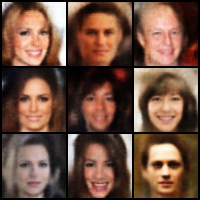}
\subcaption{Sample 4}\label{fig: celeba-sample-4}
\end{minipage}
\caption{Randomly generated samples on the CelebA-64 dataset using NegBio-VAE. Each image is generated from a different random latent variable 
$z$ under identical model settings, illustrating the NegBio-VAE’s ability to produce diverse and realistic samples.}
\label{app.celeba-sample}
\end{center}
\end{figure*}

\end{document}